\pdfoutput=1
\documentclass[11pt]{article}

\usepackage[preprint]{acl}

\usepackage[utf8]{inputenc}
\usepackage{newunicodechar}
\newunicodechar{∞}{\ensuremath{\infty}}
\usepackage{times}
\usepackage{latexsym}
\usepackage[edges]{forest}
\usepackage[T1]{fontenc}
\usepackage[utf8]{inputenc}
\usepackage{inconsolata}

\usepackage{amsmath}
\usepackage{booktabs}
\usepackage{multirow}
\usepackage{graphicx}
\usepackage{caption}
\usepackage{subcaption}
\usepackage{amsfonts}
\usepackage{algorithm}
\usepackage{algorithmic}
\usepackage{longtable}
\usepackage{amssymb}
\usepackage{pifont}
\usepackage{xcolor}
\usepackage{tabularx}
\usepackage{bbding}
\usepackage{utfsym}
\usepackage{hyperref}
\usepackage{enumitem}
\usepackage{setspace}
\usepackage{courier} 
\usepackage{url}            % simple URL typesetting
\usepackage{nicefrac}       % compact symbols for 1/2, etc.
\usepackage{microtype}      % microtypography
\usepackage{xspace}
\usepackage{wrapfig}
\usepackage{graphicx}
\usepackage{lipsum}

\usepackage{array}      
\usepackage{tabularx}   
\usepackage{booktabs}   
\usepackage{colortbl} 
\usepackage{caption}  
\usepackage{hyperref} 
\usepackage{arydshln}
\usepackage{amssymb}
\usepackage{pifont}

\definecolor{hidden-red}{RGB}{205, 44, 36}
\definecolor{hidden-blue}{RGB}{194,232,247}
\definecolor{hidden-orange}{RGB}{243,202,120}
\definecolor{hidden-green}{RGB}{34,139,34}
\definecolor{hidden-pink}{RGB}{255,245,247}
\definecolor{hidden-black}{RGB}{20,68,106}

\newcommand{\eg}{e.g.\@\xspace}

\usepackage{array}
\usepackage{makecell}
\newcolumntype{P}[1]{>{\raggedright\arraybackslash\hspace{3pt}}p{#1}}

\title{Towards Agentic RAG with Deep Reasoning: \\A Survey of RAG-Reasoning Systems in LLMs}

\author{
    Yangning Li$^{1}$\thanks{$\,$ Equal Contribution.}, 
    Weizhi Zhang$^{2}$\footnotemark[1], 
    Yuyao Yang$^{2}$,
    Wei-Chieh Huang$^{2}$,
    Yaozu Wu$^{3}$\\
    \textbf{
    Junyu Luo$^{4}$, Yuanchen Bei$^{5}$, Henry Peng Zou$^{2}$, Xiao Luo$^{6}$, Yusheng Zhao$^{4}$
    } \\
    \textbf{
    Chunkit Chan$^{7}$, Yankai Chen$^{2}$, Zhongfen Deng$^{2}$, Yinghui Li$^{1}$, Hai-Tao Zheng$^{1}$, 
    } \\
    \textbf{
    Dongyuan Li$^{3}$, Renhe Jiang$^{3}$, Ming Zhang$^{4}$, Yangqiu Song$^{7}$, Philip S. Yu$^{1}$
    } \\
    $^{1}$Tsinghua University
    $^{2}$University of Illinois Chicago
    $^{3}$The University of Tokyo\\
    $^{4}$Peking University 
    $^{5}$University of Illinois Urbana-Champaign\\
    $^{6}$University of California, Los Angeles
    $^{7}$HKUST\\
\texttt{ynli23@mails.tsinghua.edu.cn}, \texttt{wzhan42@uic.edu}
}

\begin{document}

\maketitle

\begin{abstract}

Retrieval-Augmented Generation (RAG) lifts the factuality of Large Language Models (LLMs) by injecting external knowledge, yet it falls short on problems that demand multi-step inference; conversely, purely reasoning-oriented approaches often hallucinate or mis-ground facts. This survey synthesizes both strands under a unified reasoning-retrieval perspective. We first map how advanced reasoning optimizes each stage of RAG (\textbf{Reasoning-Enhanced RAG}). Then, we show how retrieved knowledge of different type supply missing premises and expand context for complex inference (\textbf{RAG-Enhanced Reasoning}). Finally, we spotlight emerging \textbf{Synergized RAG-Reasoning} frameworks, where (agentic) LLMs iteratively interleave search and reasoning to achieve state-of-the-art performance across knowledge-intensive benchmarks. We categorize methods, datasets, and open challenges, and outline research avenues toward deeper RAG-Reasoning systems that are more effective, multimodally-adaptive, trustworthy, and human-centric. The collection is available at \url{https://github.com/DavidZWZ/Awesome-RAG-Reasoning}.

\end{abstract}
\section{Introduction}

\begin{figure*}[t]
    \centering
    \includegraphics[width=2\columnwidth]{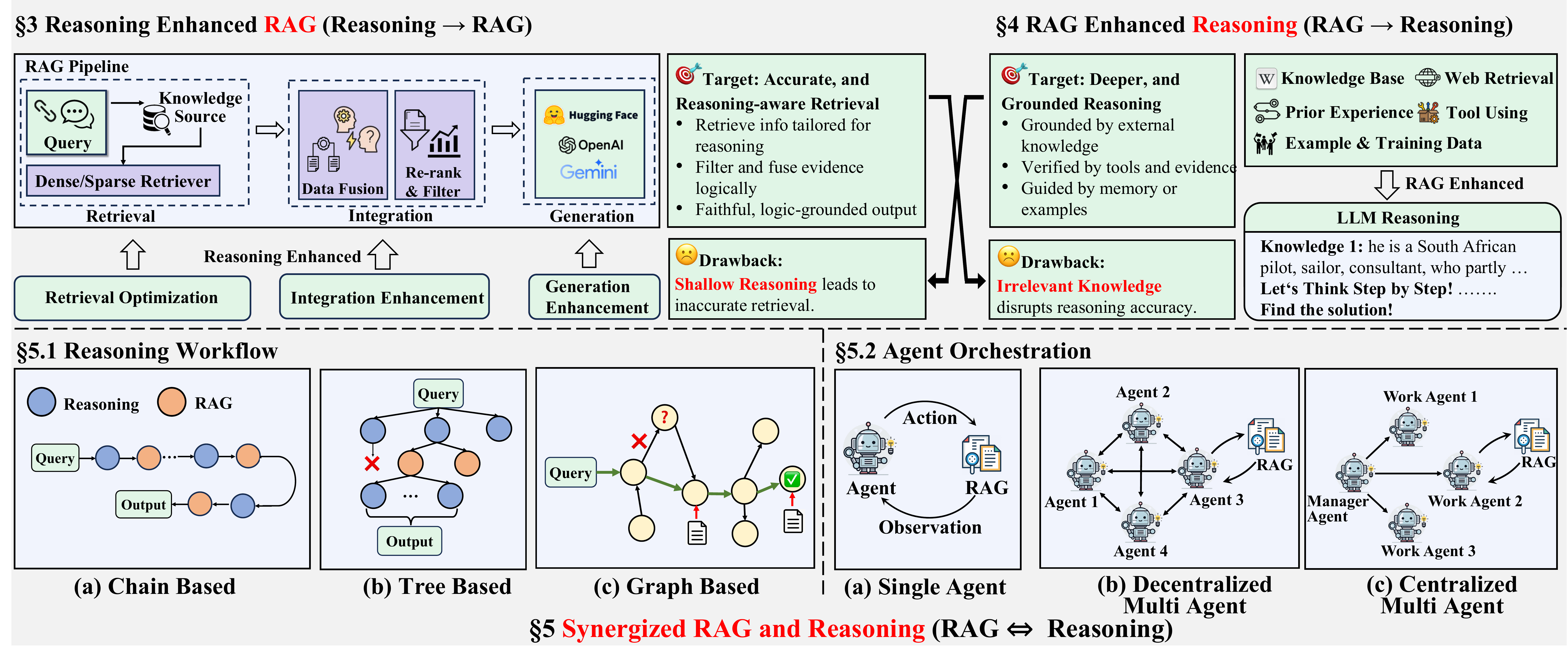}
    \caption{Overview of the RAG-Reasoning System.
The \textit{Reasoning-Enhanced RAG} methods and \textit{RAG-Enhanced Reasoning} methods represent \textbf{one-way} enhancements. In contrast, the \textit{Synergized RAG-Reasoning System} performs reasoning and retrieval {\textbf{iteratively}}, enabling mutual enhancements.}
    \label{fig:overview}
\end{figure*}
\vspace{-5pt}

The remarkable progress in Large Language Models (LLMs) has transformed a wide array of fields, showcasing unprecedented capabilities across diverse tasks \citep{zhao2023survey}. Despite these advancements, the effectiveness of LLMs remains hindered by two fundamental limitations: {knowledge hallucinations}, due to the static and parametric manner of their knowledge storage \citep{huang2025survey}; and {struggles with complex reasoning}, especially when tackling real-world problems \citep{chang2024survey}. These limitations have driven the development of two major directions: Retrieval-Augmented Generation (RAG) \citep{fan2024survey}, which provides LLMs with external knowledge; and various methods aimed at enhancing their inherent reasoning abilities \citep{chen2025towards}.

The two limitations are inherently intertwined: missing knowledge can impede reasoning, and flawed reasoning hinders knowledge utilization \citep{tonmoy2024comprehensive}. Naturally, researchers have increasingly explored combining retrieval with reasoning, though early work followed \textit{two separate, one-way enhancements}.
The first, \textbf{Reasoning-enhanced RAG} \citep{gao2023retrieval} (Reasoning $\rightarrow$ RAG), leverages reasoning to improve specific stages of the RAG pipeline.
The second path, \textbf{RAG-enhanced Reasoning} \citep{fan2024survey} (RAG $\rightarrow$ Reasoning), supplies external factual grounding or contextual cues to bolster LLM reasoning.

While beneficial, the above methods remain bound to a static Retrieval-Then-Reasoning (RTR) framework, offering only localized improvements to individual components. Several inherent limitations persist: \textit{{(1) Retrieval Adequacy and Accuracy}} cannot be guaranteed; Pre-retrieved knowledge may fail to align with the actual knowledge needs that emerge during reasoning, especially in complex tasks \citep{zheng2025deepresearcher,li2025benchmarking}. \textit{{(2) Reasoning Depth}} remains constrained. When retrieved knowledge contains errors or conflicts, it can adversely interfere with the model's inherent reasoning capabilities \citep{li2025search,chen2025research}. {\textit{(3) System Adaptability}} proves insufficient. The RTR framework lacks mechanisms for iterative feedback or dynamic retrieval during reasoning. This rigidity limits its effectiveness in scenarios that require adaptive reasoning, such as open-domain QA or scientific discovery \citep{xiong2025rag,alzubi2025open}.

As shown in Figure \ref{fig:overview}, these shortcomings have catalyzed a paradigm shift toward \textbf{Synergized Retrieval and Reasoning} within LLMs (RAG $\Leftrightarrow$ Reasoning). These methods support a dynamic, iterative interplay where reasoning actively guides retrieval, and newly retrieved knowledge, in turn, continuously refines the reasoning process.
This trend is further exemplified by recent ''Deep Research'' products from OpenAI\footnote{\url{https://openai.com/index/introducing-deep-research/}}, Gemini\footnote{\url{https://gemini.google/overview/deep-research/}}, Perplexity\footnote{\url{https://www.perplexity.ai/hub/blog/introducing-perplexity-deep-research}}, and others, which emphasize tightly coupled retrieval and reasoning \cite{zhang2025web}. These systems employ agentic capabilities to orchestrate multi-step web search and leverage reasoning to comprehensively interpret retrieved content, solving problems demanding in-depth investigation.

This survey charts the shift from isolated enhancements to cutting-edge synergized frameworks where retrieval and reasoning are deeply interwoven and co-evolve. While surveys on RAG \citep{fan2024survey,gao2023retrieval} and LLM Reasoning \citep{chen2025towards,li2025system} exist, a dedicated synthesis focusing on their integration remains lacking. Our goal is to provide a comprehensive overview of how the symbiosis between retrieval and reasoning is advancing LLM capabilities, with particular emphasis on the move towards a synergized RAG and Reasoning framework.

The survey is structured as follows: Section \ref{sec:pre} introduces the background; Section \ref{sec: Reason4RAG} and \ref{sec: RAG-enhanced Reasoning} review two one-way enhancements, respectively. Section \ref{sec: RAReason} unifies both lines into synergized RAG–Reasoning frameworks. Section \ref{sec: benchmarks} lists benchmarks, and Section \ref{sec:future} outlines open challenges.

\definecolor{hidden-draw}{RGB}{0,0,0}
\definecolor{ReasoningenhancedRAGClr}{RGB}{230,210,255}
\colorlet {ReasoningenhancedRAGLight}{ReasoningenhancedRAGClr!55}

\definecolor{RAGenhancedReasoningClr}{RGB}{190,215,255}
\colorlet {RAGenhancedReasoningLight}{RAGenhancedReasoningClr!55}

\definecolor{SynergizedRAGandReasoningClr}{RGB}{255,196,141}
\colorlet {SynergizedRAGandReasoningLight}{SynergizedRAGandReasoningClr!55}

\tikzstyle{my-box}=[
        rectangle,
        draw=hidden-draw,
        rounded corners,
        text opacity=1,
        minimum height=1.5em,
        minimum width=5em,
        inner sep=2pt,
        align=center,
        fill opacity=.5,
        line width=0.8pt,
]

\tikzstyle{leaf}=[
        my-box, 
        minimum height=1.5em, 
        fill=white, text=black, align=left,
        font=\tiny,
        inner xsep=2pt,
        inner ysep=4pt,
        line width=0.8pt,
]

\begin{figure*}[t]
    % \vspace{-2mm}
    \centering
    \resizebox{\textwidth}{!}{
        
        \begin{forest}
            forked edges,
            for tree={
                grow=east,
                reversed=true,
                anchor=base west,
                parent anchor=east,
                child anchor=west,
                base=left,
                font=\footnotesize,
                anchor=center,
                align=center,
                text centered,
                rectangle,
                draw=hidden-draw,
                rounded corners,
                align=left,
                minimum width=2em,
                edge+={darkgray, line width=1pt},
                s sep=3pt,
                inner xsep=2pt,
                inner ysep=3pt,
                line width=0.8pt,
                ver/.style={rotate=90, child anchor=north, parent anchor=south,anchor=center,font=\LARGE\bfseries},
            },
            [
                A survey of RAG and Reasoning, ver
                [
                    Reasoning-enhanced\\
                    RAG ~(\S\ref{sec: Reason4RAG}), fill= ReasoningenhancedRAGClr, text width= 9em, anchor=center, align=center, font=\normalsize
                    [
                        Retrieval  \\Optimization ~(\S\ref{sec:re_2_rag_ret}), fill=ReasoningenhancedRAGLight, text width= 10em, anchor=center, align=center, font=\normalsize
                        [
                            Reasoning-Aware Query Reformulation~(\S\ref{sec:ret_query}) , fill=ReasoningenhancedRAGLight, text width= 20em, anchor=center, align=center, font=\normalsize
                            [   
                                \eg~
                                 Collab-RAG \citep{xu2025collab}{,} DynQR \citep{anonymous2025dynqr}{,} DeepRetrieval \citep{jiang2025deepretrieval}, 
                                leaf, fill=ReasoningenhancedRAGLight, text width=44em, anchor=center, align=center, text centered, font = \normalsize
                            ]
                        ]
                        [
                            Retrieval Strategy and Planning~(\S\ref{sec:ret_strategy}), fill=ReasoningenhancedRAGLight, text width= 20em, anchor=center, align=center, font=\normalsize
                            [
                                \eg~
                                PAR-RAG \citep{zhang2025credible}{,} LPKG \citep{wang2024learning}{,} FIND \citep{jia2025find}
                                , leaf, fill=ReasoningenhancedRAGLight, text width=44em, anchor=center, align=center, text centered, font = \normalsize
                            ]
                        ]
                        [
                            Retrieval Model Enhancement~(\S\ref{sec:ret_model}), fill=ReasoningenhancedRAGLight, text width= 20em, anchor=center, align=center, font=\normalsize
                            [
                                \eg~
                                GNN-RAG \citep{mavromatis2024gnn}{,} RuleRAG \citep{chen2024rulerag}{,} 
                                , leaf, fill=ReasoningenhancedRAGLight, text width=44em, anchor=center, align=center, text centered, font = \normalsize
                            ]
                        ]
                    ]
                    [
                        Integration \\Enhancement~(\S\ref{sec:re_2_rag_integ}), fill=ReasoningenhancedRAGLight, text width= 10em, anchor=center, align=center, font=\normalsize
                        [
                            Relevance Assessment \& Filtering~(\S\ref{sec:integ_relev}), fill=ReasoningenhancedRAGLight, text width= 20em, anchor=center, align=center, font=\normalsize
                            [
                            \eg~
                            SEER \citep{zhao2024seer}{,} M-RAG-R \citep{yoran2024making}
                            , leaf, fill=ReasoningenhancedRAGLight, text width=44em, anchor=center, align=center, text centered, font = \normalsize
                            ]
                        ]
                        [
                            Information Synthesis \& Fusion~(\S\ref{sec:integ_info}), fill=ReasoningenhancedRAGLight, text width= 20em, anchor=center, align=center, font=\normalsize
                            [
                            \eg~
                            BeamAggR \citep{chu2024beamaggr}{,} DualRAG \citep{cheng2025dualrag} {,} CRP-RAG \citep{xu2024crp}
                            , leaf, fill=ReasoningenhancedRAGLight, text width=44em, anchor=center, align=center, text centered, font = \normalsize
                            ]
                        ]
                    ]
                    [
                        Generation  \\Enhancement~(\S\ref{sec:re_2_rag_gen}), fill=ReasoningenhancedRAGLight, text width= 10em, anchor=center, align=center, font=\normalsize
                        [
                            Context-Aware Generation~(\S\ref{sec:gen_con}), fill=ReasoningenhancedRAGLight, text width= 20em, anchor=center, align=center, font=\normalsize
                            [
                            \eg~
                            Open-RAG \citep{islam2024openrag}{,} RARE~\citep{wang2025rare}{,} Self-Reasoning \citep{xia2025improving}
                            , leaf, fill=ReasoningenhancedRAGLight, text width=44em, anchor=center, align=center, text centered, font = \normalsize
                            ]
                        ]
                        [
                            Grounded Generation Control~(\S\ref{sec:gen_ground}), fill=ReasoningenhancedRAGLight, text width= 20em, anchor=center, align=center, font=\normalsize
                            [
                            \eg~
                             RARR \citep{gao-etal-2023-rarr}{,} TRACE \citep{DBLP:conf/emnlp/FangMM24}{,} AlignRAG \citep{wei2025alignrag}
                            , leaf, fill=ReasoningenhancedRAGLight, text width=44em, anchor=center, align=center, text centered, font = \normalsize
                            ]
                        ]
                    ]
                ]
                [
                    RAG-enhanced \\
                    Reasoning~(\S\ref{sec: RAG-enhanced Reasoning}), fill= RAGenhancedReasoningClr, text width= 9em, anchor=center, align=center, font=\normalsize
                    [
                        External Knowledge\\Retrieval~(\S\ref{sec: External Knowledge Retrieval}), fill=RAGenhancedReasoningLight, text width= 10em, anchor=center, align=center, font=\normalsize
                        [
                            Knowledge Base (\S\ref{sec: Knowledge Base}), fill=RAGenhancedReasoningLight, text width= 19em, anchor=center, align=center, font=\normalsize
                            [
                            \eg~
                            Premise-Retrieval \cite{tao2025assisting}{, }
                            ReaRAG \cite{lee2025rearag}{, }
                            CBR-RAG \cite{wiratunga2024cbr}
                            % PIKE-RAG \cite{wang2025pike}
                            % LeanDojo~\cite{yang2023leandojo}{;}
                            % SubgraphRAG \cite{li2024simple}
                            , leaf, fill=RAGenhancedReasoningLight, text width=45em, anchor=center, align=center, text centered, font = \normalsize
                            ]
                        ]
                        [
                            Web Retrieval (\S\ref{sec: Web Retrieval}), fill=RAGenhancedReasoningLight, text width= 19em, anchor=center, align=center, font=\normalsize
                            [
                            \eg~
                            ALR$^2$ \cite{li2024alr} {, }
                            % FRVA~\cite{fan2024frva}{;}
                            RARE \cite{tran2024rare}{, }
                            % SSFV~\cite{vladika2025step}{;}
                            Open-RAG~\cite{islam2024openrag}
                            % RAG-Star \cite{jiang2024rag}
                            , leaf, fill=RAGenhancedReasoningLight, text width=45em, anchor=center, align=center, text centered, font = \normalsize
                            ]
                        ]
                        [
                            Tool Using (\S\ref{sec: Tool Using}), fill=RAGenhancedReasoningLight, text width= 19em, anchor=center, align=center, font=\normalsize
                            [
                            \eg~
                            % LATM~\cite{cai2024largelanguagemodelstool}{;}
                            % MORE~\cite{cui-etal-2024-multi}{;}
                            % AVATAR \cite{wu2024avatar}{;}
                            TATU \cite{li2024towards}{, }
                            TRICE\cite{qiao2024making}{, }
                            Re-Invoke \cite{chen2024re}
                            % ToolLLM \cite{qin2023toolllm}
                            % Meta-Reasoning \cite{alazraki2025metareasoningimprovestooluse}
                            , leaf, fill=RAGenhancedReasoningLight, text width=45em, anchor=center, align=center, text centered, font = \normalsize
                            ]
                        ]
                    ]
                    [
                        In-Context\\Retrieval~(\S\ref{sec: In-context Retrieval}), fill=RAGenhancedReasoningLight, text width= 10em, anchor=center, align=center, font=\normalsize
                        [
                            Prior Experience (\S\ref{sec: Prior Experience}), fill=RAGenhancedReasoningLight, text width= 19em, anchor=center, align=center, font=\normalsize
                            [
                            \eg~
                            % RAHL \cite{sun2024retrieval}{;}
                            RAP \cite{kagaya2024rap}{, }
                            % CoPS \cite{yang2024cops}{;}
                            % HeadKV \cite{fu2024headsmatterheadlevelkv}{;}\\
                            JARVIS-1 \cite{wang2024jarvis}{, }
                            EM-LLM \cite{fountas2024human}
                            , leaf, fill=RAGenhancedReasoningLight, text width=45em, anchor=center, align=center, text centered, font = \normalsize
                            ]
                        ]
                        [
                            Example or Training Data (\S\ref{sec: Example or Training Data}), fill=RAGenhancedReasoningLight, text width= 19em, anchor=center, align=center, font=\normalsize
                            [
                            \eg~
                            MoD \cite{wang2024mixture}{, }
                            % LLM-R~\cite{wang-etal-2024-learning}{;}
                            % Dr.ICL \cite{luo2023dr}{;}
                            RE4 \cite{li2024recall}{, }
                            UPRISE \cite{cheng2023uprise}
                            % OpenRAG \cite{zhou2025openrag}
                            , leaf, fill=RAGenhancedReasoningLight, text width=45em, anchor=center, align=center, text centered, font = \normalsize
                            ]
                        ]
                    ]
                ]
                [
                Synergized RAG-\\ 
                Reasoning ~(\S\ref{sec: RAReason}), fill= SynergizedRAGandReasoningClr, text width= 9em, anchor=center, align=center, font=\normalsize
                    [
                        Reasoning Workflow \\ (\S\ref{sec:workflow}), fill=SynergizedRAGandReasoningLight, text width= 10em, anchor=center, align=center, font=\normalsize
                        [
                            Chain-based (\S\ref{sec:chain}), fill=SynergizedRAGandReasoningLight, text width= 13.5em, anchor=center, align=center, font=\normalsize
                            [   
                                \eg~
                                IRCoT~\cite{trivedi2023interleaving}{,}
                                Rat~\cite{wang2024rat}{,}
                                CoV-RAG~\cite{he2024retrieving}{,}
                                RAFT~\cite{zhang2024raft} 
                                , leaf, fill=SynergizedRAGandReasoningLight, text width=50.5em, anchor=center, align=center, text centered, font = \normalsize
                            ]
                        ]
                        [
                            Tree-based \\ (\S\ref{sec:tree}), fill=SynergizedRAGandReasoningLight, text width= 13.5em, anchor=center, align=center, font=\normalsize
                            [   
                                \textbf{ToT}
                                \eg~
                                RATT~\citep{zhang2025ratt}{,}
                                Tree of Clarifications~\cite{kim2023tree}{,}
                                GROVE~\cite{wen2023grove}{,}\\
                                \textbf{MCTS}
                                \eg~
                                AirRAG~\cite{feng2025airrag}{,}
                                MCTS-RAG~\cite{hu2025mcts}{,}
                                SeRTS~\cite{hu2024serts}{}
                                % CORAG~\cite{wang2024corag}\\
                                , leaf, fill=SynergizedRAGandReasoningLight, text width=50.5em, anchor=center, align=center, text centered, font = \normalsize
                            ]
                        ]
                        [
                            Graph-based \\(\S\ref{sec:graph}), fill=SynergizedRAGandReasoningLight, text width= 13.5em, anchor=center, align=center, font=\normalsize
                             [   
                                \textbf{Walk-on-Graph:}
                                \eg~
                                QA-GNN~\cite{yasunaga2021qa}{,} 
                                % GNN-RAG~\cite{mavromatis2024gnn}{,}
                                LightRAG~\cite{guo2024lightrag}{,}
                                StructRAG~\cite{li2024structrag}\\
                                \textbf{Think-on-Graph:}
                                \eg~
                                ToG~\cite{sunthink}{,}
                                ToG-2.0~\cite{ma2024think}{,}
                                Graph-CoT~\cite{jin2024graph}{,}\\
                                , leaf, fill=SynergizedRAGandReasoningLight, text width=50.5em, anchor=center, align=center, text centered, font = \normalsize
                            ]
                        ]
                    ]
                    [
                        Agent Orchestration \\(\S\ref{sec:agent}), fill=SynergizedRAGandReasoningLight, text width= 10em, anchor=center, align=center, font=\normalsize
                        [
                            Signle-Agent \\ (\S~\ref{sec:single_agent}), fill=SynergizedRAGandReasoningLight, text width= 13.5em, anchor=center, align=center, font=\normalsize
                            [
                            \textbf{Prompting:} \eg~
                            ReAct~\cite{yao2023react}{,}
                            Search-O1~\cite{li2025search}{;}
                            % DeepRAG~\cite{guan2025deeprag}{;}\\
                            \textbf{SFT:} \eg~
                            Toolformer~\cite{schick2023toolformer}{,}\\
                            INTERS~\cite{zhu2024inters}{;}
                            \textbf{RL:} \eg~
                            Search-R1~\cite{jin2025search}{}
                            R1-Searcher~\cite{song2025r1}
                            , leaf, fill=SynergizedRAGandReasoningLight, text width=50.5em, anchor=center, align=center, text centered, font = \normalsize
                            ]
                        ]
                        [
                            Multi-Agent \\  (\S~\ref{sec:multi_agent}), fill=SynergizedRAGandReasoningLight, text width= 13.5em, anchor=center, align=center, font=\normalsize
                             [   
                                \textbf{Decentralized:}
                                \eg~
                                M-RAG~\cite{wang2024m}{,} 
                                MDocAgent~\cite{han2025mdocagent}{,}
                                Agentic reasoning~\cite{wu2025agentic}\\
                                \textbf{Centralized:}
                                \eg~
                                HM-RAG~\cite{liu2025hm}{,}
                                SurgRAW~\cite{low2025surgraw}{,}
                                Chain of Agents~\cite{zhang2024chain}\\
                                , leaf, fill=SynergizedRAGandReasoningLight, text width=50.5em, anchor=center, align=center, text centered, font = \normalsize
                            ]
                        ]
                    ]
                ]
            ]
        \end{forest}
    }
    \caption{Taxonomy of Recent Advances in RAG-Reasoning System.}
    \label{fig:taxonomy}
\end{figure*}
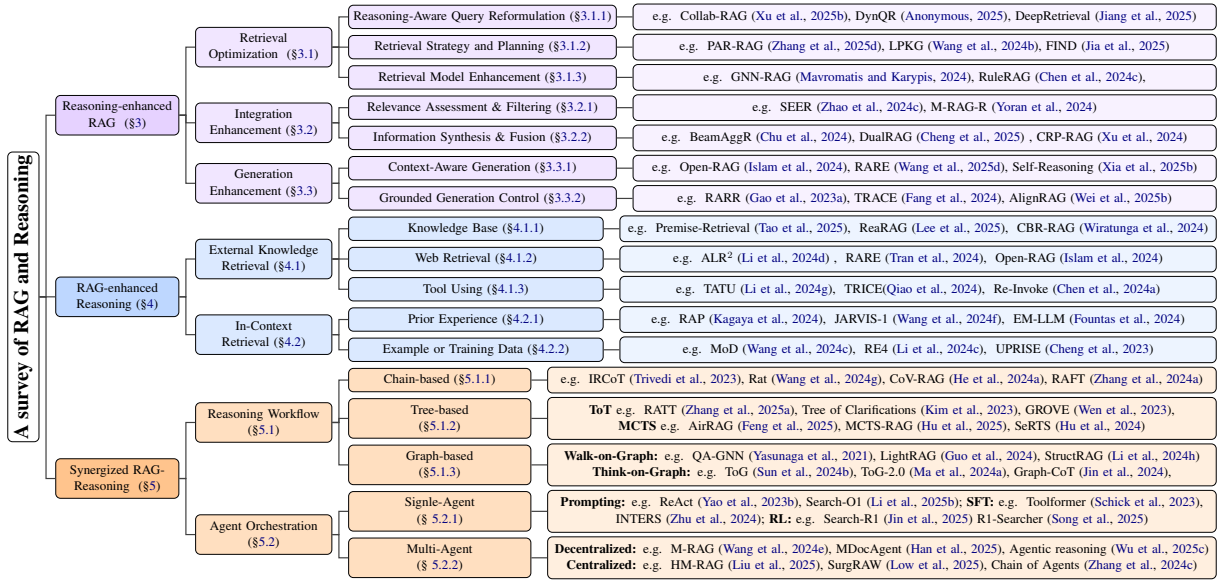

\section{Background and Preliminary}
\label{sec:pre}

RAG mitigates knowledge cut-off of LLMs through three sequential stages: \textit{(i)} \textit{Retrieval}, fetching task-relevant content from external knowledge stores;\textit{ (ii)} \textit{Integration}, deduplicating, resolving conflicts, and re-ranking the retrieved content; and \textit{(iii) Generation}, reasoning over the curated context to produce the final answer. Concurrently, Chain-of-Thought technique has significantly enhanced the reasoning capabilities of modern LLMs by encouraging them to ``\textit{think step by step}'' before answering. The synergy between the structured RAG pipeline and these multi-step reasoning capacities grounds the emerging RAG-Reasoning paradigm explored in this survey.
\section{Reasoning-Enhanced RAG} \label{sec: Reason4RAG}
Traditional RAG methods first retrieve relevant documents, then concatenate the retrieved knowledge with the original query to generate the final answer. These methods often fail to capture the deeper context or intricate relationships necessary for complex reasoning tasks. By integrating reasoning capabilities across \textbf{Retrieval}, \textbf{Integration}, and \textbf{Generation} stages of the RAG pipeline, the system can identify and fetch the most relevant information, reducing hallucinations and improving response accuracy.\footnote{If reasoning only serves to better leverage \textbf{fixed} retrieved knowledge in a unidirectional manner, it is considered within \S \ref{sec:re_2_rag_gen}. In contrast, if reasoning \textbf{dynamically triggers new retrieval}, it is discussed in \S \ref{sec: RAReason}.}

\subsection{Retrieval Optimization}
\label{sec:re_2_rag_ret}
Retrieval optimization leverages reasoning to improve result relevance and quality.
Existing methods are broadly categorized (1) Reasoning-Aware Query Reformulation, (2) Retrieval Strategy and Planning, and (3) Retrieval Model Enhancement.

\subsubsection{Reasoning-Aware Query Reformulation}\label{sec:ret_query}
It reformulates the original query to better retrieve reasoning-relevant context. 
First, {query decomposition} breaks down complex queries into simpler sub-queries~\citep{xu2025collab}. 
Second, {query reformulation} recasts ambiguous queries into more clear ones.
To align with reasoning needs of generator, certain works train rewrites with RL signals~\cite{anonymous2025dynqr,wang2025maferw}.
Third, {query expansion} enrich the semantic richness of the query via CoT reasoning~\cite{dhuliawala2024chain,li2024can,lee2024radcot}.

\subsubsection{Retrieval Strategy and Planning}\label{sec:ret_strategy}
This section covers global retrieval guidance. 
{Advance planning} uses a reasoning model to generate a complete retrieval blueprint prior to execution. PAR-RAG \citep{zhang2025credible} applies CoT for multi-step planning, mitigating local optima. LPKG \citep{wang2024learning} fine-tunes LLMs on knowledge graphs to encode relational structure. 
In contrast, {adaptive retrieval decision} methods make a one-step prediction on whether and how to retrieve. FIND \citep{jia2025find} and adaptive RAG \citep{jeong2024adaptive} use classifiers to assess query complexity and select retrieval strategies, reducing unnecessary calls. \citet{marina2025llm} further adds features like entity popularity and question type.

\subsubsection{Retrieval Model Enhancement}\label{sec:ret_model}
A line of work enhances retrievers with reasoning via two strategies. The first one {leverages structured knowledge}: GNN-RAG \citep{mavromatis2024gnn} encodes knowledge graphs with GNNs for implicit multi-hop reasoning, while RuleRAG \citep{chen2024rulerag} appends symbolic rules to guide retrieval toward logical consistency. Another strategy {integrates explicit reasoning}: \citet{ji2024retrieval} combines CoT with the query to improve intermediate knowledge recall in multi-hop QA. 

\subsection{Integration Enhancement}
\label{sec:re_2_rag_integ}

Integration enhancement uses reasoning to assess relevance and merge heterogeneous evidence, preventing irrelevant content from disrupting generation. Methods fall into two categories: (1) relevance assessment and (2) information synthesis.

\subsubsection{Relevance Assessment \& Filtering}\label{sec:integ_relev}
These methods assess the relevance of each retrieved fragment to the user query through deeper reasoning. SEER \citep{zhao2024seer} employs assessor experts to select faithful, helpful, and concise evidence while discarding irrelevant content. \citet{yoran2024making} improves robustness by filtering non-entailing passages using an NLI model, then fine-tuning the LLM on mixed relevant/irrelevant contexts to help it ignore residual noise.

\subsubsection{Information Synthesis \& Fusion}\label{sec:integ_info}
Once relevant snippets are identified, the challenge is to fuse them into a coherent evidence set. BeamAggR \citep{chu2024beamaggr} enumerates sub-question answer combinations and aggregates them via probabilistic reasoning. DualRAG \citep{cheng2025dualrag} combines reasoning-augmented querying with progressive knowledge aggregation to filter and organize retrieved information into an evolving outline. CRP-RAG \citep{xu2024crp} builds a reasoning graph to retrieve, evaluate, and aggregate knowledge at each node, dynamically selecting knowledge-sufficiency paths before generation.

\subsection{Generation Enhancement}
\label{sec:re_2_rag_gen}
Even with retrieved context, traditional RAG may still generate unfaithful content without reasoning. Reasoning during generation addresses this issue through two main approaches: (1) context-aware synthesis and (2) grounded generation control.

\subsubsection{Context-Aware Synthesis Strategies}\label{sec:gen_con}
Context-aware generation ensures outputs remain relevance while reducing noise. {Selective–context utilization} prunes or re-weights content based on task relevance.  
Open-RAG \citep{islam2024openrag} uses a sparse expert mixture to dynamically select knowledge modules, while RARE~\citep{wang2025rare} adds domain knowledge to prompts to promote reliance on external context over memorization. 
{Reasoning path generation} builds explicit logical chains to enhance transparency, e.g.,
\citet{ranaldi2024eliciting} generate contrasting explanations by comparing paragraph relevance step-by-step, guiding the model toward accurate conclusions. Self-Reasoning \citep{xia2025improving} constructs structured reasoning chains through sequential evidence selection and verification.

\subsubsection{Grounded Generation Control}\label{sec:gen_ground}
Grounded generation control introduces verification mechanisms to ensure outputs remain anchored to retrieved evidence through reasoning. 
{Fact verification} methods use reasoning to assess factual consistency between generated content and retrieved evidence, e.g., Self-RAG \citep{asai2023selfrag} introduces reflection markers during decoding to trigger critical review and correction.  
{Citation generation} links generated content to source materials to enhance traceability and credibility, as in RARR \citep{gao-etal-2023-rarr}, which inserts citations while preserving stylistic coherence. 
{Faithful reasoning} ensures that each reasoning step adheres to retrieved evidence without introducing unverified content. 
TRACE \citep{DBLP:conf/emnlp/FangMM24} builds knowledge graphs to form coherent evidence chains, while AlignRAG \citep{wei2025alignrag} applies criticism alignment to refine reasoning paths.

\section{RAG-Enhanced Reasoning} \label{sec: RAG-enhanced Reasoning}

Integrating external knowledge or in-context knowledge during reasoning can help LLMs reduce hallucinations and bridge logical gaps. External retrieval leverages structured sources like databases or web content, providing factual grounding, like IAG \cite{zhang2023iag}. In-context retrieval utilizes internal contexts like prior interactions or training examples, enhancing contextual coherence, like RA-DT \cite{schmied2024retrieval}. Both strategies collectively improve factual accuracy, interpretability, and logical consistency of reasoning processes.
 
\subsection{External Knowledge Retrieval} \label{sec: External Knowledge Retrieval}

External knowledge retrieval incorporates web content, database information, or external tools into reasoning, effectively filling knowledge gaps. Targeted retrieval improves factual accuracy, enabling language models to reliably address complex queries by grounding reasoning steps in verified external evidence.

\subsubsection{Knowledge Base} \label{sec: Knowledge Base}

Knowledge base (KB) typically stores arithmetic, commonsense, or logical knowledge in databases, books, or documents, with retrieval approaches varying by task. For {question answering (QA) reasoning}, AlignRAG \cite{wei2025alignrag}, MultiHop-RAG \cite{tang2024multihop}, and CRP-RAG \cite{xu2025crp} retrieve interconnected factual entries from general KBs to enhance sequential reasoning. In {specialized reasoning tasks}, mathematical approaches like Premise-Retrieval \cite{tao2025assisting} and ReaRAG \cite{lee2025rearag} utilize formal lemmas from theorem libraries for structured deduction; legal approaches like CASEGPT \cite{yang2024casegpt} and CBR-RAG \cite{wiratunga2024cbr} extract judicial precedents for analogical reasoning. For {code generation tasks}, CodeRAG \cite{li2025coderag} and \citet{koziolek2024llm} access code snippets from repositories, ensuring syntactic correctness.

\subsubsection{Web Retrieval} \label{sec: Web Retrieval}

Web retrieval accesses dynamic online content like web pages, news or social media. Specifically, in {fact-checking tasks}, approaches such as VeraCT Scan \cite{niu2024veract}, Ragar \cite{khaliq2024ragar}, PACAR \cite{zhao2024pacar}, and STEEL \cite{li2024re} verify claims step-by-step using evidence from news or social media, enhancing logical reasoning. Meanwhile, {QA-based reasoning} like RARE \cite{tran2024rare}, RAG-Star \cite{jiang2024rag}, MindSearch \cite{chen2024mindsearch}, and OPEN-RAG \cite{islam2024openrag} iteratively refine reasoning with broad web content, aligning with current trends in agentic search, which involve synthesizing complex online materials to enhance context-aware and robust reasoning. Conversely, in specialized areas like {medical reasoning}, approaches such as FRVA \cite{fan2024frva} and ALR$^2$ \cite{li2024alr}, retrieve literature for accurate diagnostics.

\subsubsection{Tool Using} \label{sec: Tool Using}

Tool-using approaches leverage external resources like calculators, libraries, or APIs to enhance reasoning interactively. In {QA-based reasoning}, Re-Invoke \cite{chen2024re}, AVATAR \cite{wu2024avatar}, ToolkenGPT \cite{hao2023toolkengpt}, and ToolLLM \cite{qin2023toolllm} invoke calculators or APIs (e.g., Yahoo Finance, Wikidata), improving numerical accuracy and factual precision. Within the context of {scientific modeling}, SCIAGENT \cite{ma2024sciagent} and TRICE \cite{qiao2024making} integrate symbolic computation tools (e.g., WolframAlpha), strengthening computational robustness. Similarly, in {mathematical computation}, llm-tool-use \cite{luo2025self} autonomously employs calculators for accurate numerical reasoning. Distinctively in {code generation tasks}, RAR \cite{dutta2024rar} retrieves code documentation via OSCAT libraries, ensuring syntactic accuracy and executable logic.

\subsection{In-context Retrieval} \label{sec: In-context Retrieval}

In-context retrieval leverages a model’s internal experiences or retrieved examples from demonstrations and training data to guide reasoning. This retrieval provides relevant exemplars, guiding models to emulate reasoning patterns and enhancing accuracy and logical coherence in novel questions.

\subsubsection{Prior Experience} \label{sec: Prior Experience}

Prior experience refers to past interactions or successful strategies stored in a model’s internal memory, with retrieval varying by task. In tasks involving {planning and decision-making tasks} such as robot path finding, RAHL \cite{sun2024retrieval} and RA-DT \cite{schmied2024retrieval} leverage past decisions and reinforcement signals for sequential reasoning. For {interactive reasoning tasks}, JARVIS-1 \cite{wang2024jarvis}, RAP \cite{kagaya2024rap}, and EM-LLM \cite{fountas2024human} dynamically recall multimodal interactions and conversational histories, facilitating adaptive reasoning for personalized interactions. In the domain for \textbf{logical reasoning}, CoPS \cite{yang2024cops} retrieves structured prior cases like medical and legal cases for robust logical reasoning in medical and legal scenarios.

\subsubsection{Example or Training Data} \label{sec: Example or Training Data}

Unlike approaches relying on prior experiences, example-based reasoning retrieves external examples from demonstrations or training data. For example, In {complex text-understanding}, RE4 \cite{li2024recall} and \citet{fei2024retrieval} utilize annotated sentence pairs to enhance relation recognition. Addressing {QA-based reasoning}, OpenRAG \cite{zhou2025openrag}, UPRISE \cite{cheng2023uprise}, MoD \cite{wang2024mixture}, and Dr.ICL \cite{luo2023dr} select demonstrations closely matching queries, improving generalization. Additionally, in {code generation tasks}, PERC \cite{yoo2025perc} retrieves pseudocode by semantic or structural similarity from datasets like HumanEval, ensuring alignment with target code.

\section{Synergized RAG-Reasoning} \label{sec: RAReason}

Many real-world problems, such as open-domain question answering~\cite{yang2015wikiqa, chen2020open} and scientific discovery~\cite{lu2024ai, wang2023scientific, baek2024researchagent, schmidgall2025agent}, require an iterative approach where new evidence continuously informs better reasoning and vice versa. A single retrieval step may not provide sufficient information, and a single round of reasoning may overlook key insights \cite{trivedi2023interleaving}. By tightly integrating retrieval and reasoning in a multi-step, interactive manner, these systems can progressively refine both the search relevance of retrieved information and the reasoning-based understanding of the original query.
We focus on two complementary perspectives within existing approaches: reasoning workflows, which emphasize structured, often pre-defined inference formats for multi-step reasoning; and agent orchestration, which focus on how agents interact with environment and coordinate with each others.

\subsection{Reasoning Workflow}\label{sec:workflow}
Broadly, the reasoning workflows can be categorized as chain-based, tree-based, or graph-based, reflecting an evolution from linear reasoning chains to branching and expressive reasoning structures.

\subsubsection{Chain-based}\label{sec:chain}
Chain-of-Thought (CoT)~\cite{wei2022chain} structures the reasoning process as a linear sequence of intermediate steps.
However, relying solely on the parametric knowledge of LLMs can lead to error propagation.
To solve this, IRCoT~\cite{trivedi2023interleaving} and Rat~\cite{wang2024rat} interleave retrieval operations between reasoning steps. 
Several recent methods further improve the robustness and rigor of this chain-based paradigm via verification and filtering.
CoV-RAG~\cite{he2024retrieving} introduces a chain-of-verification that checks and corrects each reasoning step against retrieved references. To combat noisy or irrelevant context, approaches like RAFT~\cite{zhang2024raft} fine-tune LLMs to ignore distractor documents, while Chain-of-Note~\cite{yu2024chain} prompts the model to take sequential “reading notes” on retrieved documents to filter out unhelpful information.

\subsubsection{Tree-based}\label{sec:tree}
Tree-based reasoning methods typically adopt either Tree-of-Thought (ToT)~\cite{yao2023tree} or Monte Carlo Tree Search (MCTS)~\cite{browne2012survey} approaches.
\textbf{ToT} extends the CoT to explicitly construct a deterministic reasoning tree and branch multiple logical pathways.
Examples include RATT~\citep{zhang2025ratt}, which construct retrieval-augmented thought trees to simultaneously evaluate multiple reasoning trajectories.
Such ToT principles avoid LLM being trapped by an early mistaken assumption and have been applied to address ambiguous questions~\cite{kim2023tree}, to cover different diagnostic possibilities~\cite{yang2025tree}, and to create complex stories~\cite{wen2023grove}.
Conversely, \textbf{MCTS}-based approaches like AirRAG~\cite{feng2025airrag}, MCTS-RAG~\cite{hu2025mcts}, and SeRTS~\cite{hu2024serts} employ probabilistic tree search, dynamically prioritizing exploration based on heuristic probabilities. To ensure retrieval and reasoning quality, AirRAG~\citep{feng2025airrag} incorporates self-consistency checks, and MCTS-RAG~\cite{hu2025mcts} integrates adaptive MCTS retrieval to refine evidence and reduce hallucinations. 

\subsubsection{Graph-based}\label{sec:graph}

\textbf{Walk-on-Graph} methods mainly rely on graph learning techniques for the retrieval and reasoning. 
For example, PullNet~\cite{sun2019pullnet}, QA-GNN~\cite{yasunaga2021qa}, 
% GNN-RAG \cite{mavromatis2024gnn}, 
and GreaseLM~\cite{zhanggreaselm} directly integrate graph neural networks (GNNs) to iteratively aggregate information from neighbor nodes, excelling at modeling the intricate relationships inherent in graph-structured data.
Methods such as SR~\cite{zhang2022subgraph}, LightRAG~\cite{guo2024lightrag}, and StructRAG~\cite{li2024structrag} employ lightweight graph techniques such as vector indexing and PageRank to efficiently retrieve and reason in multi-hop context, providing the LLM with high-quality, structured content tailored for the queries. 
In contrast, \textbf{Think-on-Graph} methods integrate graph structures directly into the LLM reasoning loop, enabling dynamic and iterative retrieval and reasoning processes guided by the LLMs themselves.
In the Think-on-Graph (ToG) framework \cite{sunthink, ma2024think}, the LLM uses the KG as a “reasoning playground”: at each step, it decides which connected entity or relation to explore next, gradually building a path that leads to the answer. 
While Graph‑CoT \cite{jin2024graph} introduces a three‑stage iterative loop (reasoning, graph interaction, and execution), KGP \cite{wang2024knowledge} prioritize first constructing a document‑level KG, both enabling LLM‑driven graph traversal agent to navigate passages in each step with globally coherent context. GraphReader \cite{li2024graphreader} further refines this paradigm by coupling LLM reasoning with explicit subgraph retrieval and evidence anchoring at each step
% Similarly, Graph-CoT \cite{jin2024graph}, Knowledge Graph Prompting (KGP) \cite{wang2024knowledge} and GraphReader \cite{li2024graphreader} engage LLMs in iterative reasoning loops explicitly guided by graph interactions, ensuring each reasoning step with corresponding evidence. 

\subsection{Agent Orchestration}\label{sec:agent}

According to agent architectures \cite{luo2025large}, we organize existing work into single-agent and multi-agent. Particularly, we have attached recent advances in agentic deep research and implementations in Appendix~\ref{sec:appendix_deep_research}.

\subsubsection{Single-Agent}\label{sec:single_agent}

Single agentic system interweaves knowledge retrieval (search) into an LLM’s reasoning loop, enabling dynamic information lookup at each step of problem solving and incentivizing it to actively seek out relevant evidence when needed.

The ReAct~\cite{yao2023react} paradigm and its derivatives~\cite{li2025search, alzubi2025open} have pioneered this \textbf{prompting} strategy by guiding LLMs to explicitly alternate between reasoning steps and external tool interactions, such as database searches. 
Different from ReAct that separates reasoning and action, with explicit commands like ``search'' triggering external retrieval, methods such as Self-Ask~\cite{press2023measuring} and IRCoT~\cite{trivedi2023interleaving} prompt the model to recursively formulate and answer sub-questions, enabling interleaved retrieval within the Chain-of-Thought (step-by-step retrieval and reasoning). 
Involving self-reflection strategies, DeepRAG~\cite{guan2025deeprag} and Self-RAG~\cite{asai2024selfrag} empower LLMs to introspectively assess their knowledge limitations and retrieve only when necessary.

Rather than relying solely on prompting or static retrievers, Toolformer~\cite{schick2023toolformer} and INTERS~\cite{zhu2024inters} represent a complementary approach via \textbf{supervised fine-tuning} (SFT) LLMs on instruction-based or synthetic datasets that interleave search and reasoning.
Synthetic data generation~\cite{schick2023toolformer, mao2024rag, zhang2024raft} aims to create large-scale, diverse, and task-specific datasets for search without the need for extensive human annotation. 
In contrast, instruction-based data reformulation~\cite{zhu2024inters, wang2024instructretro, lin2023ra, nguyen2024sfr} repurposes existing datasets into instructional formats to fine-tune models for improved generalization and alignment with human-like reasoning. 
INTERS~\cite{zhu2024inters} exemplifies this approach by introducing a SFT dataset encompassing 20 tasks, derived from 43 distinct datasets with manually written templates.

\textbf{Reinforcement learning} 
(RL)-incentivized approaches provides a mechanism to optimize answer quality via reward signals on incentivizing agents’ behaviors – what to search, how to integrate retrieved evidence, and when to stop, aiming at complex knowledge-intensive tasks (or “deep research” questions). 
Notable efforts like WebGPT~\cite{nakano2021webgpt} and RAG-RL~\cite{huang2025rag} focus on improving reasoning fidelity by rewarding outputs based on factual correctness or human preference. 
More recent contributions operate directly in dynamic environments (e.g., live web search, local search tools), training agents to explore, reflect, and self-correct in noisy real-world conditions. For example, Search‑R1~\cite{jin2025search} learns to generate <search> token during reasoning and concurrently R1‑Searcher~\cite{song2025r1} builds on RL-driven search demonstrating strong generalization across domains. Deep-Researche~\cite{zheng2025deepresearcher} make step further by introducing the first end-to-end RL-trained research agent that interacts with the open web.
These settings showcase emergent capabilities, like decomposition, iterative verification, and retrieval planning, that supervised methods often hard to instill.
Moreover, ReSearch~\cite{chen2025learning} and ReARTeR~\cite{sun2025rearter} tackle a deeper challenge: not just producing correct answers, but aligning reasoning steps with both factuality and interpretability.

\subsubsection{Multi-Agent}\label{sec:multi_agent}

The exploration of multi-agent collaboration within RAG and reasoning has led to diverse orchestrations: centralized architectures (harness collective intelligence from workers-manager paradigm) and decentralized architectures (leverage complementary capabilities from role-specialized agents).

\textbf{Decentralized architectures} deploy multiple agents to collaboratively perform retrieval, reasoning, and knowledge integration, aiming to broaden coverage of relevant information and fully exploit the heterogeneous strengths of specialized agents. 
\citet{wang2024m} and \citet{salve2024collaborative} introduce multi-agent systems where each agent retrieves from a partitioned database or a specific data source (relational databases, NoSQL document stores, etc.). 
Beyond retrieval, Collab-RAG~\cite{xu2025collab} and RAG-KG-IL~\cite{yu2025rag} integrate different model capacities and assign them different roles in reasoning and knowledge integration. 
This philosophy extends to multi-modal settings as in MDocAgent~\cite{han2025mdocagent}, which employs a team of text and image agents to process and reason the document-based QA.
A general formulation is seen in Agentic reasoning~\cite{wu2025agentic}, which unites tool-using agents for search, computation, and structured reasoning, orchestrated to solve complex analytical tasks.

\textbf{Centralized architectures} structure agents in hierarchical centralized patterns, supporting efficient task decomposition and progressive refinement. HM-RAG~\cite{liu2025hm} and SurgRAW~\cite{low2025surgraw} both employ decomposer-retriever-decider architectures, where different agent roles isolate subproblems such as multimodal processing or surgical decision-making. \citet{wu2025talk} and \citet{iannelli2024sla} emphasize dynamic routing and system reconfiguration, respectively—enabling intelligent agent selection based on task relevance or resource constraints. Chain of Agents~\cite{zhang2024chain} and the cooperative multi-agent control framework for on-ramp merging~\cite{zhang2025cascading} illustrate hierarchical agent designs where layered processing enables long-context summarization or policy refinement. Collectively, these works demonstrate how centralized control and hierarchical pipelining foster efficiency and adaptability in multi-agent RAG-reasoning systems.

\section{Benchmarks and Datasets} 
\label{sec: benchmarks}

% In your preamble, ensure xcolor is loaded with table and dvipsnames:
% \usepackage[table,dvipsnames]{xcolor}

\renewcommand{\arraystretch}{1.4}

% === Original color definitions ===
\definecolor{header}{RGB}{217,225,242}
\definecolor{rowA}{RGB}{248,248,248}
% Single-hop QA (teal tones)
\definecolor{shqDark}{RGB}{150,200,200}
\definecolor{shqLight}{RGB}{200,230,230}
% Multi-hop QA
\definecolor{mhqDark}{RGB}{141,180,226}
\definecolor{mhqLight}{RGB}{198,219,239}
% Multi-choice QA
\definecolor{mcqDark}{RGB}{215,186,218}
\definecolor{mcqLight}{RGB}{235,222,236}
% Text Summarization
\definecolor{sumDark}{RGB}{253,203,110}
\definecolor{sumLight}{RGB}{255,228,166}
% Dialog / Math
\definecolor{dlgDark}{RGB}{158,202,163}
\definecolor{dlgLight}{RGB}{207,232,198}
% Graph QA
\definecolor{gqaDark}{RGB}{204,170,128}
\definecolor{gqaLight}{RGB}{232,208,184}
% Fact Checking
\definecolor{fchDark}{RGB}{255,153,51}
\definecolor{fchLight}{RGB}{255,229,204}
% Code
\definecolor{codeDark}{RGB}{170,170,170}
\definecolor{codeLight}{RGB}{230,230,230}
% Web Browsing
\definecolor{wbDark}{RGB}{255,182,193}
\definecolor{wbLight}{RGB}{255,228,225}

% === Define text‐font colors as a darker mix of each task’s dark shade and black ===
\colorlet{wbFont}{wbDark!80!black}
\colorlet{shqFont}{shqDark!80!black}
\colorlet{mhqFont}{mhqDark!80!black}
\colorlet{mcqFont}{mcqDark!80!black}
\colorlet{sumFont}{sumDark!80!black}
\colorlet{dlgFont}{dlgDark!80!black}
\colorlet{gqaFont}{gqaDark!80!black}
\colorlet{fchFont}{fchDark!80!black}
\colorlet{codeFont}{codeDark!80!black}

% === Column types & spacing ===
\newcolumntype{T}[1]{>{\bfseries\raggedright\arraybackslash}p{#1}} % Task (bold)
\newcolumntype{D}[1]{>{\itshape\raggedright\arraybackslash}p{#1}} % Dataset (italic)
\newcolumntype{C}[1]{>{\centering\arraybackslash}p{#1}}           % Centered
\setlength{\tabcolsep}{0.1pt}

\begin{table*}[htpb]
  \centering
  \fontsize{6pt}{6pt}\selectfont
    \scalebox{0.95}{
  \begin{tabularx}{\textwidth}{@{}%
    T{1.6cm}  % Task
    D{3.1 cm}  % Dataset
    C{1.1 cm} C{2.0cm} C{2.3cm} C{1.6cm}
    C{1.0cm} C{1.6cm} C{1.6cm}
  @{}}
    \toprule
    \rowcolor{header}
    \textbf{Task} &
    \textbf{Dataset} & \textbf{Domain} & \textbf{Knowledge Source} &
    \textbf{Knowledge Type} & \textbf{Reasoning} &
    \textbf{Size} & \textbf{Input} & \textbf{Output} \\
    \midrule

    % Web Browsing
    \rowcolor{wbLight}
    \textcolor{wbFont}{Web Browsing}    & \textit{BrowseComp} \citep{wei2025browsecomp}       & General   & Human, Internet            & Commonsense, Logical & Deductive            & 1,266     & Question/Text & Natural Language \\
    \rowcolor{wbLight}
    \textcolor{wbFont}{}                 & \textit{GAIA} \citep{mialon2023gaia}    & General   & Internet, TooL                 & Commonsense, Logical & Deductive            & 466       & Question/Text, Image/File/Code & Natural Language \\
    \rowcolor{wbLight}
    \textcolor{wbFont}{}                 & \textit{WebWalkerQA} \citep{wu2025webwalker}    & General   & Human, LLM                 & Commonsense, Logical & Deductive            & 680       & Question/Text & Natural Language \\
    \midrule

    % Single-hop QA
    \rowcolor{shqLight}
    \textcolor{shqFont}{Single-hop QA}    & \textit{TriviaQA} \citep{joshi2017triviaqa}        & General   & Internet                   & Commonsense, Logical & Deductive            & 650,000+  & Question/Text & Natural Language \\
    \rowcolor{shqLight}
    \textcolor{shqFont}{}                 & \textit{NQ} \citep{kwiatkowskinatural}              & General   & Internet                   & Commonsense, Logical & Deductive            & 307,373   & Question/Text & Natural Language \\
    \midrule

    % Multi-hop QA
    \rowcolor{mhqLight}
    \textcolor{mhqFont}{Multi-hop QA}     & \textit{2WikiMultiHopQA} \citep{ho2020constructing}  & General   & Internet                   & Commonsense, Logical & Deductive            & 192,606   & Question/Text & Natural Language \\
    \rowcolor{mhqLight}
    \textcolor{mhqFont}{}                 & \textit{HotpotQA} \citep{yang2018hotpotqa}       & General   & Internet                   & Commonsense           & Deductive            & 113,000   & Question/Text & Natural Language \\
    \rowcolor{mhqLight}
    \textcolor{mhqFont}{}                 & \textit{MuSiQue} \citep{trivedi2022musique}        & General   & Previous Resource, Internet   & Commonsense, Logical & Deductive            & 25,000    & Question/Text & Natural Language \\
    \midrule

    % Multi-choice QA
    \rowcolor{mcqLight}
    \textcolor{mcqFont}{Multi-choice QA}  & \textit{QuALITY} \citep{pang2022quality}          & Narrative & Books                      & Commonsense, Logical & Deductive, Abductive & 6,737     & Question/Text, Options & Options \\
    \rowcolor{mcqLight}
    \textcolor{mcqFont}{}                 & \textit{MMLU-Pro} \citep{wang2025mmlu}        & Science   & Previous Resource, Internet   & Arithmetic, Commonsense, Logical & Deductive, Inductive & 12,032 & Question/Text, Options & Natural Langue, Number, Options \\
    \midrule

    % Text Summarization
    % \rowcolor{sumLight}
    % \textcolor{sumFont}{Text Summarization} & \textit{XSum} \citep{narayan2018don}           & Narrative & Internet, Media            & Logical, Commonsense & Abductive           & 226,711   & Question/Text & Natural Language \\
    % \rowcolor{sumLight}
    % \textcolor{sumFont}{}                  & \textit{BIGPATENT} \citep{sharma2019bigpatent}      & Patent    & Internet                   & Commonsense, Logical & Abductive           & 1.3\,M    & Question/Text & Natural Language \\
    % \midrule

    % Dialog / Math
    \rowcolor{dlgLight}
    \textcolor{dlgFont}{Math}             & \textit{MATH} \citep{hendrycks2measuring}            & Math      & Exam                       & Arithmetic, Logic     & Deductive           & 12,500    & Question/Text, Figure, Equation & Natural Langue, Number \\
    \rowcolor{dlgLight}
    \textcolor{dlgFont}{}                  & \textit{AQuA} \citep{ling2017program}            & Math      & Exam, Internet, Previous Resource & Arithmetic, Logic & Deductive           & 100,000   & Question/Text, Options, Equation & Natural Langue, Options \\
    \midrule

    % Graph QA
    % \rowcolor{gqaLight}
    % \textcolor{gqaFont}{Graph QA}         & \textit{GraphQA} \citep{he2024g}         & General   & Previous Resource             & Commonsense, Multimodal & Deductive, Abductive & 107,503  & Question/Text & Natural Language \\
    % \rowcolor{gqaLight}
    % \textcolor{gqaFont}{}                  & \textit{GRBENCH} \citep{jin2024graph}         & General   & LLM, Human                & Logical                & Deductive, Inductive & 1,740     & Question/Text & Natural Language \\
    % \midrule

    % Fact Checking
    % \rowcolor{fchLight}
    % \textcolor{fchFont}{Fact Checking}     & \textit{Fever} \citep{thorne2018fever}            & General   & Internet                   & Logical                & Deductive, Abductive & 185,445  & Question/Text, Links & Natural Langue, Null, Options \\
    % \rowcolor{fchLight}
    % \textcolor{fchFont}{}                  & \textit{PubHealth} \citep{kotonya2020explainable}      & Health    & Internet                   & Commonsense, Logical   & Abductive, Deductive & 11,800   & Question/Text & Natural Langue, Options \\
    % \midrule

    % Code
    \rowcolor{codeLight}
    \textcolor{codeFont}{Code}            & \textit{Refactoring Oracle} \citep{tsantalis2020refactoringminer} & Software & Internet, Human         & Logical                & Deductive            & 7,226    & Code, Instruction & Code \\
    \rowcolor{codeLight}
    \textcolor{codeFont}{}                 & \textit{LiveCodeBench} \citep{jain2024livecodebench}   & Contest   & Internet                   & Logical                & Deductive, Abductive & 500+      & Question/Text, Code, Instruction & Code, Test Output \\

    \bottomrule
  \end{tabularx}}
  \caption{Overview of representative knowledge and reasoning intensive benchmarks by task category.}
  \label{tab:rag_by_task}
  \vspace{-0.4cm}
\end{table*}
\vspace{-5pt}
Benchmarks and datasets for simultaneously evaluating knowledge (RAG) and reasoning capability cover a wide range of complexities, from basic fact retrieval to intricate multi-step reasoning in general or specific domains. We categorize notable benchmarks in several tasks and list them in Table~\ref{tab:rag_by_task} and highlight their details and properties. These representative tasks include Web browsing, such as BrowseComp \citep{wei2025browsecomp}, single-hop QA, such as TriviaQA \citep{joshi2017triviaqa}, multi-hop QA, such as HotpotQA \citep{yang2018hotpotqa}, multiple-choice QA, such as MMLU-Pro \citep{wang2025mmlu}, mathematics, such as MATH \citep{hendrycks2measuring}, and code-centric evaluations from LiveCodeBench \citep{jain2024livecodebench}. More tasks can refer to Appendix ~\ref{sec:appendix_fullbenchnark} and Table ~\ref{tab:rag_benchmarks}.

\section{Future Work}
\label{sec:future}

Future research directions for Synergized RAG-Reasoning systems center around enhancing both reasoning and retrieval capabilities to meet real-world demands for accuracy, efficiency, trust, and user alignment. We outline several key challenges and opportunities below.

\begin{itemize}[fullwidth]
  \item \textbf{Reasoning Efficiency.} Despite their advantages in complex reasoning, Synergized RAG-Reasoning systems can suffer significant latency due to iterative retrieval and multi-step reasoning loops \citep{sui2025stop}. For instance, executing a single deep research query can take over 10 minutes in practical settings. This issue is especially pronounced in chain-based workflows discussed in Section~\ref{sec: RAReason}. Future research should explore reasoning efficiency through latent reasoning approaches and strategic control over reasoning depth via thought distillation and length-penalty \cite{xia2025tokenskip,zhang2025lightthinker}. Beyond reasoning itself, emerging directions in models compression like quantization, pruning, and knowledge distillation is worth to explore for efficient small RAG-reasoning systems. 
  
  \item \textbf{Retrieval Efficiency.} On the retrieval side, efficiency demands budget-aware query planning and memory-aware mechanisms that cache prior evidence or belief states to reduce redundant access \citep{zhao2024retrieval}. Additionally, adaptive retrieval control, learning when and how much to retrieve based on uncertainty signals can reduce wasteful operations. These technical paths push the system beyond static RAG, toward dynamic self-regulation of efficient retreival behaviors under real-world constraints.
  \item \textbf{Human-Agent Collaboration.} Many applications of RAG-Reasoning, such as literature reviews or interactive programming, are inherently personalized and cannot assume users know precisely what to ask or how to process retrieved results \cite{sun2025symbioticrag}. Corresponding to Section~\ref{sec:agent}, humans can act as advanced agents, providing nuanced feedback to steer reasoning processes. Future systems should develop methods for modeling user intent under uncertainty~\cite{zhang2025cold, yang2025cold}, building interactive interfaces for iterative clarification, and designing agents that adapt reasoning strategies based on user expertise and preferences \citep{zhang2025personaagent}. This human-in-the-loop approach~\cite{zou2025survey} is essential for creating robust and user-aligned RAG-Reasoning systems in open-ended domains.
  \item \textbf{Agentic Structures and Capabilities.} A key feature of Synergized RAG-Reasoning is its agentic architecture, where the system autonomously decides the roles of different agents and which tools or retrieval strategies to invoke during inference stages \citep{luo2025large, bei2025graphs}. To fully exploit this potential, future research should focus on developing agent frameworks capable of dynamic tool selection, retrieval planning, and adaptive orchestration across reasoning workflows. Such capabilities enable flexible, context-aware problem solving and are critical for handling diverse, complex tasks \citep{schneider2025generative}.
\end{itemize}

\begin{itemize}[fullwidth]
\item \textbf{Multimodal Retrieval.} As also shown in our benchmark analysis, most existing Synergized RAG-Reasoning systems remain confined to text-only tasks. However, real-world applications increasingly require the ability to retrieve and integrate multimodal content \citep{liang2024survey}. Future research should move beyond the traditional vision-text paradigm to achieve genuine multimodality. This advancement necessitates strengthening foundational abilities of MLLMs, including grounding and cross-modal reasoning \citep{liang2024survey}. Additionally, enhancing the agentic capabilities of these models through hybrid-modal chain-of-thought reasoning is crucial, enabling interaction with the real world via multimodal search tools \citep{wang2025multimodal}. Concurrently, developing unified multimodal retrievers that can jointly embed images, tables, text, and heterogeneous documents is essential.

\item \textbf{Retrieval Trustworthiness.} Synergized RAG-Reasoning systems remain vulnerable to adversarial attacks through poisoned or misleading external knowledge sources. Ensuring the trustworthiness of retrieved content is therefore crucial for maintaining fully reliable downstream reasoning \citep{huang2024survey}. Techniques like watermarking and digital fingerprinting have been employed to enhance system traceability. However, there's a pressing need to develop more dynamic and adaptive methods that can keep pace with the evolving landscape of LLMs, emerging attack techniques, and shifting model contexts \citep{liu2024survey}. Existing studies have also individually explored uncertainty quantification and robust generation to bolster system reliability \citep{shorinwa2025survey}. Future research should aim to integrate these approaches, as their combination can mutually reinforce system robustness and trustworthiness. Moreover, future efforts should also focus on extending current benchmarks to encompass multi-dimensional trust metrics beyond mere accuracy.

\end{itemize}

\section{Conclusion}

This survey charts the rapid convergence of retrieval and reasoning in LLMs. We reviewed three evolutionary stages: (1) Reasoning-Enhanced RAG, which uses multi-step reasoning to refine each stage of RAG; (2) RAG-Enhanced Reasoning, which leverages retrieved knowledge to bridge factual gaps during long CoT; and (3) Synergized RAG-Reasoning systems, where single- or multi-agents iteratively refine both search and reasoning, exemplified by recent ``Deep Research'' platforms. Collectively, these lines demonstrate that tight retrieval–reasoning coupling improves factual grounding, logical coherence, and adaptability beyond one-way enhancement. Looking forward, we identify research avenues toward synergized RAG-Reasoning systems that are more effective, multimodally-adaptive, trustworthy, and human-centric.

\section*{Limitations}

While this survey synthesizes over 200 research papers across RAG and reasoning with large language models, its scope favors breadth over depth. In striving to provide a unified and comprehensive taxonomy, we may not delve deeply into the technical nuances or implementation details of individual methods-especially within specialized subfields of either RAG (e.g., sparse vs. dense retrieval, memory-augmented retrievers) or reasoning (e.g., formal logic solvers, symbolic methods, or long-context reasoning). Moreover, our categorization framework (reasoning-enhanced RAG, RAG-enhanced reasoning, and synergized RAG and reasoning) abstracts across diverse methodologies. While this facilitates a high-level understanding of design patterns, it may obscure the finer-grained trade-offs, assumptions, and limitations unique to each class of approach.

\bibliography{custom}

\clearpage
\appendix
\label{sec:appendix}
\section{Full Benchmark}
\label{sec:appendix_fullbenchnark}

Section~\ref{sec: benchmarks} introduces representative benchmarks for different RAG-reasoning tasks. This appendix complements that discussion with a comprehensive list of benchmarks organized by task and domain. Table~\ref{tab:rag_benchmarks} details each benchmark’s attributes, including the publication venue, code repository, task category, domain, primary knowledge sources, knowledge type, and reasoning capabilities. By consolidating these attributes into a single table, we facilitate the selection and comparison of benchmarks, enabling researchers to identify the most suitable datasets for future studies on RAG-enhanced reasoning.

Our benchmark compilation is primarily derived from the methods surveyed in Sections~\ref{sec: Reason4RAG} to~\ref{sec: RAReason} of this paper, with a particular focus on synergized approaches discussed in Section~\ref{sec: RAReason}. We deliberately targeted benchmarks that require both external knowledge retrieval and internal deep reasoning, as this dual requirement reflects real-world scenarios where models must not only access relevant information but also integrate and reason over it effectively. For example, in the QA domain, we include datasets that necessitate synthesizing evidence across multiple documents to answer questions that cannot be resolved through single-sentence retrieval. HotpotQA~\citep{yang2018hotpotqa} exemplifies this challenge, requiring reasoning across different Wikipedia articles. In coding tasks, benchmarks such as LiveCodeBench~\citep{jain2024livecodebench} and Refactoring Oracle~\citep{tsantalis2020refactoringminer} extend beyond pure algorithmic problem-solving by demanding retrieval of external code snippets and documentation. Similarly, in mathematics, benchmarks like MATH~\citep{hendrycks2measuring} and AQUA-RAT~\citep{das2024mathsensei} assess not only computational proficiency but also the retrieval of relevant theorems and formulas, testing the model’s ability to integrate external mathematical knowledge with internal reasoning processes.

% ─── Preamble definitions ───────────────────────────────────────────────────

\definecolor{header}{RGB}{217,225,242}   % header background
\definecolor{rowA}{RGB}{248,248,248}     % alternating row background
\newcolumntype{Y}[1]{>{\bfseries\raggedright\arraybackslash}p{#1}}
\newcolumntype{C}[1]{>{\centering\arraybackslash}p{#1}}
\newcolumntype{L}[1]{>{\raggedright\arraybackslash}p{#1}}
\newcolumntype{Z}[1]{>{\centering\itshape\arraybackslash}p{#1}}
\setlength{\tabcolsep}{0.1pt}
\renewcommand{\arraystretch}{1.35}

\setlength{\dashlinedash}{1pt}
\setlength{\dashlinegap}{1pt}
\setlength{\arrayrulewidth}{0.3pt}

% === Original color definitions ===
\definecolor{header}{RGB}{217,225,242}
\definecolor{rowA}{RGB}{248,248,248}
% Single-hop QA (teal tones)
\definecolor{shqDark}{RGB}{150,200,200}
\definecolor{shqLight}{RGB}{200,230,230}
% Multi-hop QA
\definecolor{mhqDark}{RGB}{141,180,226}
\definecolor{mhqLight}{RGB}{198,219,239}
% Multi-choice QA
\definecolor{mcqDark}{RGB}{215,186,218}
\definecolor{mcqLight}{RGB}{235,222,236}
% Text Summarization
\definecolor{sumDark}{RGB}{253,203,110}
\definecolor{sumLight}{RGB}{255,228,166}
% Dialog / Math
\definecolor{dlgDark}{RGB}{158,202,163}
\definecolor{dlgLight}{RGB}{207,232,198}
% Graph QA
\definecolor{gqaDark}{RGB}{204,170,128}
\definecolor{gqaLight}{RGB}{232,208,184}
% Fact Checking
\definecolor{fchDark}{RGB}{255,153,51}
\definecolor{fchLight}{RGB}{255,229,204}
% Code
\definecolor{codeDark}{RGB}{170,170,170}
\definecolor{codeLight}{RGB}{230,230,230}
% Web Browsing
\definecolor{wbDark}{RGB}{255,182,193}
\definecolor{wbLight}{RGB}{255,228,225}

% Long-form QA (Soft lavender)
\definecolor{lfqaDark}{RGB}{175,148,209}
\definecolor{lfqaLight}{RGB}{220,210,235}
% Multimodal QA (Soft pink)
\definecolor{mmqaDark}{RGB}{230,130,160}
\definecolor{mmqaLight}{RGB}{255,220,230}
% Multi-step QA (Light lime)
\definecolor{msqaDark}{RGB}{160,190,100}
\definecolor{msqaLight}{RGB}{220,235,190}
% Dialog (Peach)
\definecolor{dlgNewDark}{RGB}{255,180,130}
\definecolor{dlgNewLight}{RGB}{255,230,210}

% ──────────────────────────────────────────────────────────────────────────────

% ─── Part 1: up through Multimodal QA ─────────────────────────────────────────
\begin{table*}[htbp]
  \centering
  % \hspace*{-0.75cm}
  {
  \fontsize{6pt}{5.5pt}\selectfont
  \begin{tabularx}{17.3cm}{@{}%
    L{2.0cm}  Z{1.1cm}  C{0.9cm}  C{1.5cm}  C{1.5cm}
    C{1.8cm}  C{1.8cm}  C{2.0cm}  C{1.0cm}  C{1.8cm}  C{1.8cm}
  @{}}
    \toprule
    \rowcolor{header}
    \textbf{Dataset} & \textbf{Venue} & \textbf{Resource} & \textbf{Task} &
    \textbf{Domain} & \textbf{Knowledge Source} & \textbf{Knowledge Type} &
    \textbf{Reasoning Capability} & \textbf{Size} & \textbf{Input} & \textbf{Output} \\
    \midrule

    % Code
    \textbf{Code} \\
    \addlinespace[2pt]
    \rowcolor{codeLight}
    LiveCodeBench \citep{jain2024livecodebench}
      & Arxiv’24
      & \href{https://github.com/LiveCodeBench/LiveCodeBench}{Link}
      & Code & General & Internet & Logical
      & Deductive, Abductive
      & 1,055
      & Question/Text, Code, Instruction
      & Code Instance, Test Output       \\
      \rowcolor{codeLight}
    Refactoring Oracle \citep{tsantalis2020refactoringminer}
      & IEEE’22
      & \href{https://github.com/tsantalis/RefactoringMiner}{Link}
      & Code & Software & Internet, Human & Logical
      & Deductive
      & 7,226
      & Code, Instruction
      & Code Instance                     \\
      \rowcolor{codeLight}
    ColBench \citep{zhou2025sweet}
      & Arxiv’25
      & \href{https://huggingface.co/datasets/facebook/collaborative_agent_bench}{Link}
      & Code & Software & LLM, Human & Logical
      & Abductive, Inductive
      & 10,000+
      & Question/Text, Links/Sources, Code
      & Code Instance                    \\
    \hdashline
    \addlinespace[5pt]

    % Domain-specific QA
    \textbf{Math} \\
    \addlinespace[2pt]
    \rowcolor{dlgLight}
    MATH \citep{hendrycks2measuring}
      & NeurIPS’21
      & \href{https://github.com/hendrycks/math}{Link}
      & Domain‐specific QA & Math & Exam/Competition
      & Logical, Arithmetic
      & Deductive
      & 12,500
      & Question/Text, Equations
      & Number, Natural Language            \\
      \rowcolor{dlgLight}
    MiniF2F \citep{zheng2021minif2f}
      & ICLR’22
      & \href{https://github.com/openai/miniF2F}{Link}
      & Domain‐specific QA & Math & Exam/Competition, Books
      & Logical, Arithmetic
      & Deductive
      & 488
      & Question/Text, Equations
      & Number, Natural Language            \\
      \rowcolor{dlgLight}
    AQuA \citep{ling2017program}
      & Arxiv’17
      & \href{https://github.com/google-deepmind/AQuA}{Link}
      & Domain‐specific QA & Math & Previous Source, Exam/Competition, Internet
      & Arithmetic, Logical
      & Deductive
      & 100,000
      & Question/Text, Options, Equations
      & Natural Language, Options/Labels \\
    \hdashline
    \addlinespace[5pt]

    % Fact Checking
    \textbf{Fact Checking} \\
    \addlinespace[2pt]
    \rowcolor{fchLight}
    CRAG \citep{yang2024crag}
      & NeurIPS’24
      & \href{https://github.com/facebookresearch/CRAG/}{Link}
      & Fact Checking & General & Internet
      & Commonsense
      & Deductive, Abductive
      & 4,409
      & Question/Text
      & Natural Language                     \\
      \rowcolor{fchLight}
    CREAK \citep{onoe2021creak}
      & NeurIPS’21
      & \href{https://github.com/yasumasaonoe/creak}{Link}
      & Fact Checking & General & Human
      & Commonsense
      & Deductive, Abductive, Analogical
      & 13,000
      & Question/Text
      & Options/Labels, Natural Language    \\
      \rowcolor{fchLight}
    Fever \citep{thorne2018fever}
      & ACL’18
      & \href{https://github.com/awslabs/fever}{Link}
      & Fact Checking & General & Internet
      & Logical
      & Deductive, Abductive
      & 185,445
      & Question/Text, Links/Sources
      & Natural Language, Options/Labels \\
      \rowcolor{fchLight}
    PubHealth \citep{kotonya2020explainable}
      & EMNLP’20
      & \href{https://github.com/neemakot/Health-Fact-Checking}{Link}
      & Fact Checking & Health & Internet
      & Commonsense, Logical
      & Abductive, Deductive
      & 11,800
      & Question/Text
      & Natural Language, Options            \\
    \hdashline
    \addlinespace[5pt]

    % Graph QA
    \textbf{Graph QA} \\
    \addlinespace[2pt]
    \rowcolor{gqaLight}
    GraphQA \citep{he2024g}
      & NeurIPS’24
      & \href{https://github.com/XiaoxinHe/G-Retriever}{Link}
      & Graph QA & General & Previous Source
      & Commonsense, Multimodal
      & Deductive, Abductive
      & 107,503
      & Question/Text
      & Natural Language                     \\
      \rowcolor{gqaLight}
    GRBENCH \citep{jin2024graph}
      & ACL’24
      & \href{https://github.com/PeterGriffinJin/Graph-CoT}{Link}
      & Graph QA & General & LLM, Human
      & Logical
      & Deductive, Inductive
      & 1,740
      & Question/Text
      & Natural Language                     \\
    \hdashline
    \addlinespace[5pt]

    % Long-form QA
    \textbf{Long-form QA} \\
    \addlinespace[2pt]
    \rowcolor{lfqaLight}
    $\infty$ BENCH \citep{zhang2024bench}
      & Arxiv’24
      & \href{https://github.com/OpenBMB/InfiniteBench}{Link}
      & Long-form QA & General & Internet, Human
      & Multimodal, Logical
      & Inductive, Abductive
      & 3,946
      & Question/Text, Code, Equations
      & Natural Language, Number, Code Instance \\
    \hdashline
    \addlinespace[5pt]

    % Multimodal QA
    \textbf{Multimodal QA} \\
    \addlinespace[2pt]
    \rowcolor{mmqaLight}
    CrisisMMD \citep{alam2018crisismmd}
      & Arxiv’18
      & \href{https://huggingface.co/datasets/QCRI/CrisisMMD}{Link}
      & Multimodal QA & Crisis Response & Media, Internet
      & Commonsense, Multimodal
      & Abductive
      & 16,097
      & Question/Text, Figure/Image
      & Options, Natural Language            \\
      \rowcolor{mmqaLight}
    ALFWORLD \citep{shridharalfworld}
      & ICLR’21
      & \href{https://github.com/alfworld/alfworld}{Link}
      & Multimodal QA & Game & Previous Source
      & Multimodal
      & Deductive, Abductive
      & 3,827
      & Question/Text, Figure/Image
      & Natural Language                     \\
      \rowcolor{mmqaLight}
    MMLongBench-DOC \citep{ma2025mmlongbench}
      & NeurIPS’24
      & \href{https://github.com/mayubo2333/MMLongBench-Doc}{Link}
      & Multimodal QA & Narrative & Previous Source, Internet
      & Multimodal
      & Deductive, Abductive
      & 1,082
      & Figure/Image, Question/Text, Documents
      & Natural Language, Number              \\
      \rowcolor{mmqaLight}
    LongDocURL \citep{deng2024longdocurl}
      & Arxiv’24
      & \href{https://github.com/dengc2023/LongDocURL}{Link}
      & Multimodal QA & Narrative & Internet, Previous Source, LLM
      & Multimodal
      & Deductive, Abductive
      & 2,325
      & Figure/Image, Question/Text, Documents
      & Natural Language, Number              \\
      \rowcolor{mmqaLight}
    UDA \citep{huiuda}
      & NIPS'24
      & \href{https://github.com/qinchuanhui/UDA-Benchmark}{Link}
      & Multimodal QA & Narrative & Internet, Paper/Report
      & Multimodal
      & Deductive
      & 29,590
      & Documents, Question/Text
      & Natural Language, Number            \\
      \rowcolor{mmqaLight}
    SCIENCEQA \citep{lu2022learn}
      & NeurIPS’22
      & \href{https://huggingface.co/datasets/derek-thomas/ScienceQA}{Link}
      & Multimodal QA & Science & Human
      & Logical, Multimodal
      & Deductive
      & 21,000
      & Question/Text, Options, Figure/Image
      & Options, Natural Language, Number      \\
      \rowcolor{mmqaLight}
    WebShop \citep{yao2022webshop}
      & NeurIPS’22
      & \href{https://github.com/princeton-nlp/WebShop}{Link}
      & Multimodal QA & E-commerce & Internet
      & Multimodal
      & Inductive, Abductive
      & 12,087
      & Instruction, Question/Text
      & Natural Language, Image/Figure         \\
      \rowcolor{mmqaLight}
    SurgCoTBench \citep{low2025surgraw}
      & Arxiv’25
      & — 
      & Multimodal QA & Health & Human
      & Multimodal, Logical
      & Abductive, Deductive
      & 14,176
      & Question/Text, Figure/Image, Options
      & Options, Natural Language, Number      \\
    \bottomrule
  \end{tabularx}
  }
  % \captionsetup{width=18cm}
  % \captionsetup[table]{justification=raggedright,singlelinecheck=false}
  \caption{Full representative knowledge and reasoning intensive benchmarks across diverse task categories (Part 1).}
  \label{tab:rag_benchmarks}
\end{table*}

% ─── Part 2: continuation, no new label ────────────────────────────────────────
\begin{table*}[htbp]
  \centering
  % \hspace*{-0.75cm}
  {
  \fontsize{6pt}{5.5pt}\selectfont
  \begin{tabularx}{17.3cm}{@{}%
    L{2.0cm}  Z{1.1cm}  C{0.9cm}  C{1.5cm}  C{1.5cm}
    C{1.8cm}  C{1.8cm}  C{2.0cm}  C{1.0cm}  C{1.8cm}  C{1.8cm}
  @{}}
    \toprule
    \rowcolor{header}
    \textbf{Dataset} & \textbf{Venue} & \textbf{Resource} & \textbf{Task} &
    \textbf{Domain} & \textbf{Knowledge Source} & \textbf{Knowledge Type} &
    \textbf{Reasoning Capability} & \textbf{Size} & \textbf{Input} & \textbf{Output} \\
    \midrule

    % Multi-choice OA
    \textbf{Multi-choice QA} \\
    \addlinespace[2pt]
    \rowcolor{mcqLight}
    Bamboogle \citep{press2023measuring}
      & EMNLP’23
      & \href{https://github.com/ofirpress/self-ask}{Link}
      & Multi-choice QA & General & Internet
      & Logical
      & Deductive, Abductive
      & 125
      & Question/Text
      & Natural Language                     \\
      \rowcolor{mcqLight}
    BIG-Bench \citep{srivastava2022beyond}
      & Arxiv’22
      & \href{https://github.com/suzgunmirac/BIG-Bench-Hard}{Link}
      & Multi-choice QA & General & Internet
      & Commonsense, Logical
      & Deductive, Abductive, Inductive, Analogical
      & 204
      & Question/Text, Options
      & Natural Language, Number, Options/Labels \\
      \rowcolor{mcqLight}
    ADQA \citep{li2024dalk}
      & EMNLP’24
      & \href{https://github.com/David-Li0406/DALK}{Link}
      & Multi-choice QA & Health & Previous Source
      & Commonsense, Logical
      & Deductive, Abductive
      & 446
      & Question/Text, Options
      & Options                               \\
      \rowcolor{mcqLight}
    QuALITY \citep{pang2022quality}
      & NAACL’22
      & \href{https://github.com/nyu-mll/quality}{Link}
      & Multi-choice QA & Narrative & Books
      & Commonsense, Logical
      & Deductive, Abductive
      & 6,737
      & Question/Text, Options
      & Options                               \\
      \rowcolor{mcqLight}
    MMLU-Pro \citep{wang2025mmlu}
      & NeurIPS’24
      & \href{https://github.com/TIGER-AI-Lab/MMLU-Pro}{Link}
      & Multi-choice QA & Science & Previous Source, Internet
      & Arithmetic, Commonsense, Logical
      & Deductive, Inductive
      & 12,032
      & Question/Text, Options
      & Natural Language, Number, Options    \\
    \hdashline
    \addlinespace[5pt]
    % Multi-hop QA
    \textbf{Multi-hop QA} \\
    \addlinespace[2pt]
    \rowcolor{mhqLight}
    FRAMES \citep{krishna2024fact}
      & Arxiv’24
      & \href{https://huggingface.co/datasets/google/frames-benchmark}{Link}
      & Multi-hop QA & General & Internet
      & Commonsense, Logical, Arithmetic
      & Deductive
      & 824
      & Question/Text
      & Natural Language                     \\
      \rowcolor{mhqLight}
    HotpotQA \citep{yang2018hotpotqa}
      & EMNLP’18
      & \href{https://hotpotqa.github.io/}{Link}
      & Multi-hop QA & General & Internet
      & Commonsense
      & Deductive
      & 113,000
      & Question/Text
      & Natural Language                     \\
      \rowcolor{mhqLight}
    GPQA \citep{rein2024gpqa}
      & Arxiv'24
      & \href{https://github.com/idavidrein/gpqa}{Link}
      & Multi-hop QA & Science & Human
      & Logical
      & Deductive, Abductive
      & 448
      & Question/Text, Options
      & Natural Language, Number, Options    \\
      \rowcolor{mhqLight}
    HLE \citep{phan2025humanity}
      & Arxiv'25
      & \href{https://github.com/centerforaisafety/hle}{Link}
      & Multi-hop QA & Science & Human
      & Logical, Arithmetic, Multimodal
      & Deductive, Abductive
      & 2,500
      & Question/Text, Options, Figure/Image
      & Natural Language, Number, Options    \\
      \rowcolor{mhqLight}
    CWQ \citep{talmor2018web}
      & NAACL’18
      & \href{https://paperswithcode.com/paper/the-web-as-a-knowledge-base-for-answering}{Link}
      & Multi-hop QA & General & Internet
      & Commonsense
      & Deductive
      & 34,689
      & Question/Text
      & Natural Language                     \\
      \rowcolor{mhqLight}
    IIRC \citep{ferguson2020iirc}
      & EMNLP’20
      & \href{https://www.semanticscholar.org/paper/IIRC\%3A-A-Dataset-of-Incomplete-Information-Reading-Ferguson-Gardner/01a1f2df34d947d7aa5698ca6fb31c03d15a5183}{Link}
      & Multi-hop QA & General & Internet
      & Commonsense, Logical
      & Deductive
      & 13,000+
      & Question/Text, Links/Sources
      & Number, Natural Language             \\
      \rowcolor{mhqLight}

    MINTQA \citep{he2024mintqa}
      & Arxiv’24
      & \href{https://github.com/probe2/multi-hop/}{Link}
      & Multi-hop QA & General & Internet
      & Commonsense, Logical
      & Deductive
      & 10,479
      & Question/Text
      & Natural Language                     \\
      \rowcolor{mhqLight}
    MuSiQue \citep{trivedi2022musique}
      & ACL’22
      & \href{https://github.com/stonybrooknlp/musique}{Link}
      & Multi-hop QA & General & Previous Source, Internet
      & Commonsense, Logical
      & Deductive
      & 25,000
      & Question/Text
      & Natural Language                     \\
      \rowcolor{mhqLight}

    TopiOCQA \citep{adlakha2022topiocqa}
      & TACL’22
      & \href{https://github.com/McGill-NLP/topiocqa}{Link}
      & Multi-hop QA & General & Internet
      & Commonsense, Logical
      & Deductive
      & 54,494
      & Question/Text
      & Natural Language                     \\
      \rowcolor{mhqLight}
    2WikiMultiHopQA \citep{ho2020constructing}
      & COLING’20
      & \href{https://huggingface.co/datasets/xanhho/2WikiMultihopQA}{Link}
      & Multi-hop QA & General & Internet
      & Commonsense, Logical
      & Deductive
      & 192,606
      & Question/Text
      & Natural Language                     \\
    \hdashline
    \addlinespace[5pt]
    % Multi-step QA
    \textbf{Multi-step QA} \\
    \addlinespace[2pt]
    \rowcolor{msqaLight}
    StrategyQA \citep{geva2021did}
      & TACL’21
      & \href{https://paperswithcode.com/dataset/strategyqa}{Link}
      & Multi-step QA & General & Internet
      & Commonsense, Logical
      & Deductive
      & 2,780
      & Question/Text
      & Natural Language                     \\
    \hdashline
    \addlinespace[5pt]
    % Single-hop QA
    \textbf{Single-hop QA} \\
    \addlinespace[2pt]
    \rowcolor{shqLight}
    SimpleQA \citep{wei2024measuring}
      & Arxiv’24
      & \href{https://github.com/openai/simple-evals}{Link}
      & Single-hop QA & General & LLM, Human
      & Commonsense
      & Deductive
      & 4,326
      & Question/Text
      & Natural Language                     \\
      \rowcolor{shqLight}
    TriviaQA \citep{joshi2017triviaqa}
      & ACL’17
      & \href{https://nlp.cs.washington.edu/triviaqa/}{Link}
      & Single-hop QA & General & Internet
      & Commonsense, Logical
      & Deductive
      & 650,000+
      & Question/Text
      & Natural Language                     \\
      \rowcolor{shqLight}
    NQ \citep{kwiatkowskinatural}
      & ACL’19
      & \href{https://github.com/google-research-datasets/natural-questions}{Link}
      & Single-hop QA & General & Internet
      & Commonsense, Logical
      & Deductive
      & 307,373
      & Question/Text
      & Natural Language                     \\
    \hdashline
    \addlinespace[5pt]

    % Text Summarization
    \textbf{Text Summarization} \\
    \addlinespace[2pt]
    \rowcolor{sumLight}
    XSum \citep{narayan2018don}
      & EMNLP’18
      & \href{https://github.com/EdinburghNLP/XSum}{Link}
      & Text Summarization & Narrative & Internet, Media
      & Logical, Commonsense
      & Abductive
      & 226,711
      & Question/Text
      & Natural Language                     \\
      \rowcolor{sumLight}
    BIGPATENT \citep{sharma2019bigpatent}
      & ACL’19
      & \href{https://github.com/evasharma/bigpatent}{Link}
      & Text Summarization & Patent & Internet
      & Commonsense, Logical
      & Abductive
      & 1.3\,M
      & Question/Text
      & Natural Language                     \\
    \hdashline
    \addlinespace[5pt]

    % Web Browsing
    \textbf{Web Browsing} \\
    \addlinespace[2pt]
    \rowcolor{wbLight}
    BrowseComp \citep{wei2025browsecomp}
      & Arxiv'25
      & \href{https://github.com/openai/simple-evals}{Link}
      & Web Browsing & General & Human, Internet
      & Commonsense, Logical
      & Deductive
      & 1,266
      & Question/Text
      & Natural Language                     \\
      \rowcolor{wbLight}
    BrowseComp-ZH \citep{zhou2025browsecomp}
      & Arxiv'25
      & \href{https://github.com/PALIN2018/BrowseComp-ZH}{Link}
      & Web Browsing & General & Human, Internet
      & Commonsense, Logical
      & Deductive
      & 289
      & Question/Text
      & Natural Language                     \\
      \rowcolor{wbLight}
             GAIA \citep{mialon2023gaia}  & ICLR'23
      & \href{https://huggingface.co/gaia-benchmark}{Link} & Web Browsing  & General   & Internet, TooL                 & Commonsense, Logical & Deductive            & 466       & Question/Text, Image/File/Code & Natural Language \\
      \rowcolor{wbLight}
    WebWalkerQA \citep{wu2025webwalker}
      & Arxiv'25
      & \href{https://github.com/Alibaba-NLP/WebWalker}{Link}
      & Web Browsing & General & Human, LLM
      & Commonsense, Logical
      & Deductive
      & 680
      & Question/Text
      & Natural Language                     \\
        \hdashline
    \addlinespace[5pt]

    % Dialog
    \textbf{Dialog} \\
    \addlinespace[2pt]
    \rowcolor{dlgNewLight}
    DailyDialog \citep{li2017dailydialog}
      & Arxiv'17
      & \href{https://paperswithcode.com/dataset/dailydialog}{Link}
      & Dialog & General & Internet
      & Commonsense, Logical
      & –  
      & 13,118
      & Question/Text
      & Natural Language                     \\
    \bottomrule
  \end{tabularx}
  }
  % \captionsetup{width=18cm}
  % \captionsetup[table]{justification=raggedright,singlelinecheck=false}
  \caption{Full epresentative knowledge and reasoning intensive benchmarks across diverse task categories (Part 2, continued).}
\end{table*}

In addition to established benchmarks, we have incorporated newer and more challenging datasets that better mirror real-world applications. These datasets often demand extensive retrieval processes combined with expert-level or domain-specific reasoning, as seen in Humanity’s Last Exam (HLE)~\citep{phan2025humanity} and web search evaluation tasks like BrowseComp~\citep{wei2025browsecomp}. Overall, our collection encompasses \textbf{46} benchmarks covering \textbf{13} distinct tasks across \textbf{12} domains, each explicitly annotated with features such as knowledge source, knowledge type, and reasoning capacity. This breadth ensures coverage of diverse domains and task types, forming a solid foundation for evaluating the interplay between retrieval and reasoning in RAG systems.

\renewcommand{\arraystretch}{1.4}

\definecolor{header}{RGB}{217,225,242}
\definecolor{rowshade}{RGB}{245,247,250} 
\newcolumntype{T}[1]{>{\bfseries\raggedright\arraybackslash}p{#1}} 
\newcolumntype{X}{>{\raggedright\arraybackslash}X}            
\newcolumntype{C}[1]{>{\centering\arraybackslash}p{#1}}       
\setlength{\tabcolsep}{3 pt}

\rowcolors{2}{rowshade}{white}

\begin{table*}[!t]
  \centering
  \fontsize{7pt}{7pt}\selectfont
  \begin{tabularx}{\textwidth}{@{} T{3cm} C{2 cm} X X @{} }
    \toprule
    \rowcolor{header}
    \textbf{Benchmark} & \textbf{Domain} & \textbf{Primary Retrieval Challenge} & \textbf{Primary Reasoning Challenge} \\
    \midrule
    TriviaQA, NQ
      & General
      & \textbf{Scale \& Noise:} Retrieval from massive, noisy corpora.
      & \textbf{Ambiguity:} Handling real-world queries that are often underspecified or ambiguous. \\ [8pt]
    
    HotpotQA, 2WikiMultiHopQA, MuSiQue, HLE
      & General
      & \textbf{Multi-document / High-dependency Synthesis:} Requires finding and connecting evidence scattered across multiple Wikipedia articles.
      & \textbf{Multi-hop Deduction:} Explicitly designed to test the ability to link two or more discrete facts into a coherent reasoning path. \\ [8pt]
    
    MMLU-Pro, QUALITY
      & Science, Narrative
      & \textbf{Expert-level Retrieval:} Requires accessing deep specialized knowledge from academic or densely written narrative sources.
      & \textbf{Complex \& Long-form Reasoning:} MMLU-Pro demands expert-level problem-solving over rote memorization. QUALITY uniquely requires comprehension of very long texts (often >5,000 tokens). \\ [8pt]
    
    MATH, AQUA-RAT
      & Math
      & \textbf{Formal Knowledge Retrieval:} Locating precise mathematical theorems, lemmas, or formulas in formal corpora.
      & \textbf{Symbolic \& Deductive Reasoning:} Involves performing precise, multi-step logical and algebraic operations where each step must be correct. AQUA-RAT is unique in providing natural language rationales, thus testing the model's ability to explain its formal reasoning. \\ [8pt]
    
    LiveCodeBench
      & Code
      & \textbf{Structural \& Modal Heterogeneity:} Must retrieve from diverse, heterogeneous sources such as code repositories, documentation, and community forums like Stack Overflow.
      & \textbf{Tool Use \& Self-correction Reasoning:} Requires applying retrieved code snippets/APIs, executing code, and reasoning based on test outputs to debug and iteratively improve solutions. \\ [8pt]
    
    BrowseComp, WebWalkerQA
      & General (Web)
      & \textbf{Dynamism, Interactivity, and Long-tail Retrieval:} Tests agentic planning and tool use in live, unstructured web environments. BrowseComp requires creative, persistent navigation to locate hard-to-find, intertwined information, while WebWalkerQA focuses on systematic traversal of a website's subpages.
      & \textbf{Agentic \& Strategic Reasoning:} Requires planning and executing multi-step strategies (e.g., searching, clicking, extracting) in dynamic and unpredictable contexts to achieve a defined goal. \\ \noalign{\vskip 2pt}
    \bottomrule
  \end{tabularx}
  \caption{The primary retrieval and reasoning challenges for different RAG-Reasoning benchmarks.}
  \label{tab:rag_benchmarks_challenge}
\end{table*}

\rowcolors{1}{white}{white}

Within this benchmark set, single-hop QA datasets like TriviaQA~\citep{joshi2017triviaqa} focus on precise retrieval and fact recall, requiring models to locate and synthesize a single piece of evidence. In contrast, multi-hop QA benchmarks such as HotpotQA~\citep{yang2018hotpotqa} and MuSiQue~\citep{trivedi2022musique} challenge models to chain information from multiple documents and employ deductive reasoning to bridge disparate facts into coherent answers. Structured knowledge benchmarks, such as GraphQA~\citep{he2024g}, require reasoning over relational graph representations, integrating nodes and edges to resolve complex queries beyond plain text retrieval. Complementing these open-ended tasks, multiple-choice evaluations like MMLU-Pro~\citep{wang2025mmlu} test domain-specific knowledge in areas such as science, history, or law, assessing the model’s ability to perform various reasoning styles, including inductive and abductive inference. Multimodal QA benchmarks, like WebShop~\citep{yao2022webshop}, test a model’s capacity to align textual and visual information to determine the correct answer. Long-form QA datasets such as $\infty$BENCH~\citep{zhang2024bench} evaluate models’ ability to maintain logical consistency and perform inductive reasoning over lengthy contexts. Collectively, these benchmarks establish a comprehensive evaluation chain for systematically assessing RAG-reasoning capabilities.

Beyond text-based QA, RAG-augmented benchmarks span diverse tasks involving long-form generation, interactive reasoning, and domain-specific challenges in mathematics and programming. Mathematics benchmarks such as MATH~\citep{hendrycks2measuring} draw from competition-level problems to assess arithmetic and symbolic reasoning. Summarization tasks like XSum~\citep{narayan2018don} evaluate a model’s ability to condense entire news articles into concise summaries while preserving factual correctness. Fact-checking benchmarks, such as FEVER~\citep{thorne2018fever}, test the capacity for evidence retrieval and claim verification. Code-focused evaluations, including LiveCodeBench~\citep{jain2024livecodebench}, examine deductive and abductive reasoning in the context of algorithmic problem-solving. Web-based tasks, exemplified by BrowseComp~\citep{wei2025browsecomp}, emulate real-world search behavior, requiring iterative query formulation and navigation across multiple webpages.

In addition to cataloging datasets, Table~\ref{tab:rag_benchmarks_challenge} provides a synthesized overview of the primary retrieval and reasoning challenges associated with each benchmark discussed in this survey. This comparative analysis reveals critical gaps in current benchmark coverage that future research must address. From a \textbf{domain perspective}, most benchmarks still focus on a limited set of general or academic scenarios, with few tackling real-world, realistic industrial or vertical-domain tasks where retrieval sources might be personalized, proprietary or highly specialized. Regarding \textbf{retrieval capabilities}, existing benchmarks rarely test systems' ability to handle heterogeneous or multimodal content, nor do they systematically evaluate robustness against noisy, evolving, or conflicting information within a unified framework for trustworthiness. In terms of \textbf{reasoning capabilities}, current benchmarks primarily assess deductive reasoning, leaving underexplored more complex forms such as deep causal reasoning, counterfactual thinking, decision-oriented reasoning, or analogical reasoning in specialized domains. Moreover, there is a lack of standardized benchmarks and metrics for evaluating the entire reasoning-retrieval trajectory, including the efficiency of retrieval steps, the quality of intermediate queries, and the logical consistency of multi-step reasoning chains.

\section{Deep Research Implementations}
\label{sec:appendix_deep_research}

\newcolumntype{T}[1]{>{\bfseries\raggedright\arraybackslash}p{#1}}
\newcolumntype{X}{>{\raggedright\arraybackslash}X}
\newcolumntype{C}[1]{>{\centering\arraybackslash}p{#1}}
\setlength{\tabcolsep}{2 pt}
\definecolor{rowshade}{RGB}{245,247,250}  % very light gray

\rowcolors{2}{rowshade}{white}

\begin{table*}[!t]
  \centering
  \fontsize{6pt}{6pt}\selectfont
  \begin{tabularx}{\textwidth}{@{}%
      T{1.8 cm}  % Name
      C{2.3 cm}  % Base Model
      C{1.0 cm}    % Optimization
      C{0.8 cm}    % Reward
      C{1.5 cm}  % Search Engine 
      C{1.5 cm}    % Agent Architecture
      C{2.0 cm}  % Train Data
      C{3.1 cm}  % Evaluation Data
      C{0.8cm}  % Link
    @{} }
    \toprule
    \rowcolor{header}
      \textbf{Name}
    & \textbf{Base Model}
    & \textbf{Optimization}
    & \textbf{Reward}
    & \textbf{Retriever} 
    & \textbf{Agent Architecture}
    & \textbf{Train Data}
    & \textbf{Evaluation Data}
    & \textbf{Link} \\
    \midrule

    Agentic Reasoning \citep{wu2025agentic}
      & N/A
      & Prompting
      & N/A
      & Web Search
      & Centralized
      & N/A
      & GPQA
      & \href{https://github.com/theworldofagents/Agentic-Reasoning}{Link} \\

    gpt-researcher
      & 
      & Prompting
      & N/A
      & Web Search, Local Retrieval   
      & Centralized
      & N/A
      & N/A
      & \href{https://github.com/assafelovic/gpt-researcher}{Link} \\

    deep-searcher
      & Deepseek, , Claude, Gemini, Qwen
      & Prompting
      & N/A
      & Web Search
      & Hierarchical
      & N/A
      & N/A
      & \href{https://github.com/zilliztech/deep-searcher}{Link} \\

    Search-R1 \citep{jin2025search}
      & Qwen2.5-7B-Instruct, Qwen2.5-7B-Base, Qwen-2.5-3B-Instruct, Qwen-2.5-3B-Base
      & GRPO, PPO
      & Exact Match
      & Web Search
      & Single
      & NQ, HotpotQA
      & NQ, TriviaQA, PopQA, HotpotQA, 2WikiMultiHopQA, MuSiQue, Bamboogle
      & \href{https://github.com/PeterGriffinJin/Search-R1}{Link} \\

    ZeroSearch \citep{sun2025zerosearch}
      & Qwen2.5-3B-Base, Qwen2.5-7B-Base, Qwen2.5-7B-Instruct, Qwen2.5-3B-Instruct, LLaMA3.2-3B-Instruct, LLaMA3.2-3B-Base
      & GRPO, PPO, Reinforce
      & Exact Match
      & Web Search
      & Single
      & NQ, HotpotQA
      & NQ, TriviaQA, PopQA, HotpotQA, 2WikiMultiHopQA, MuSiQue, Bamboogle
      & \href{https://github.com/Alibaba-NLP/ZeroSearch}{Link} \\

    Webthinker \citep{li2025webthinker}
      & GPT-o1, GPT-o3, Deepseek-R1, QwQ-32B, Qwen2.5-32B-Instruct
      & DPO
      & Preference Pairs
      & Web Search
      & Single
      & SuperGPQA, WebWalkerQA, OpenThoughts, NaturalReasoning, NuminaMath
      & GPQA, GAIA, WebWalkerQA, Humanity’s Last Exam
      & \href{https://github.com/RUC-NLPIR/WebThinker}{Link} \\

    nanoDeepResearch
      & OpenAI series, Claude
      & Prompting
      & N/A
      & Web Search
      & Centralized
      & N/A
      & N/A
      & \href{https://github.com/liyuan24/nanoDeepResearch}{Link} \\

    \addlinespace
    DeerFlow
      & Qwen, 
      & Prompting
      & N/A
      & Web Search
      & Decentralized
      & N/A
      & N/A
      & \href{https://github.com/bytedance/deer-flow}{Link} \\

    \addlinespace
    deep-research
      & Deepseek, 
      & Prompting
      & N/A
      & Web Search
      & Single
      & N/A
      & N/A
      & \href{https://github.com/dzhng/deep-research}{Link} \\

    open-deep-research
      & OpenAI series, Deepseek, Claude, Gemini
      & Prompting
      & N/A
      & Web Search
      & Single
      & N/A
      & N/A
      & \href{https://github.com/btahir/open-deep-research}{Link} \\

    DeepResearcher \citep{zheng2025deepresearcher}
      & Qwen2.5-7B-Instruct
      & GRPO
      & Format
      & Web Search
      & Decentralized 
      & NQ, TQ, HotpotQA, 2WikiMultiHopQA
      & MuSiQue, Bamboogle, PopQA, NQ, TQ, HotpotQA, 2WikiMultiHopQA
      & \href{https://github.com/GAIR-NLP/DeepResearcher}{Link} \\

    R1-Searcher \citep{song2025r1}
      & Qwen2.5-7B-Base, Llama3.1-8B-Instruct
      & GRPO, Reinforce++, SFT
      & Retrieval, Format
      & Web Search, Local Retrieval
      & Single
      & HotpotQA, 2WikiMultiHopQA
      & HotpotQA, 2WikiMultiHopQA, MuSiQue, Bamboogle
      & \href{https://github.com/RUCAIBox/R1-Searcher}{Link} \\

    ReSearch \citep{chen2025research}
      & Qwen2.5-7B-Instruct, Qwen2.5-32B-Instruct
      & GRPO
      & Format, Answer
      & Web Search
      & Single
      & MuSiQue
      & HotpotQA, 2WikiMultiHopQA, MuSiQue, Bamboogle
      & \href{https://github.com/Agent-RL/ReSearch}{Link} \\

    Search-o1 \citep{li2025search}
      & QwQ-32B-Preview
      & Prompting
      & N/A
      & Web Search
      & Single
      & N/A
      & GPQA, MATH500, AMC2023, AIME2024, LiveCodeBench, Natural Questions, TriviaQA, HotpotQA, 2Wiki, MuSiQue, Bamboogle
      & \href{https://github.com/sunnynexus/Search-o1}{Link} \\

    r1-reasoning-rag
      & Deepseek
      & Prompting
      & N/A
      & Local Retrieval, Web Search
      & Single
      & N/A
      & N/A
      & \href{https://github.com/deansaco/r1-reasoning-rag}{Link} \\

    Open Deep Search \citep{alzubi2025open}
      & Llama3.1-70B, Deepseek-R1
      & Prompting
      & N/A
      & Web Search
      & Single
      & N/A
      & SimpleQA, FRAME
      & \href{https://github.com/sentient-agi/OpenDeepSearch}{Link} \\

    node-DeepResearch
      & Gemini, 
      & Prompting
      & N/A
      & Web Search
      & Single
      & N/A
      & N/A
      & \href{https://github.com/jina-ai/node-DeepResearch}{Link} \\

    deep-research
      & Gemini, OpenAI series, Deepseek, Claude, Grok
      & Prompt
      & N/A
      & Local Retrieval, Web Search
      & Single
      & N/A
      & N/A
      & \href{https://github.com/u14app/deep-research}{Link} \\
    \bottomrule
  \end{tabularx}
    \caption{Overview of deep research implementations.}
  \label{tab:agentic-frameworks}
\end{table*}

\rowcolors{1}{white}{white}
In this section, we extend the discussion of the agentic paradigm introduced in Section~\ref{sec:agent}, in which RAG systems adopt the role of active researchers who plan multistep queries, interleave retrieval with reasoning, and coordinate specialized tools or agents. These characteristics collectively define what we refer to as deep research, representing the ability of a system to autonomously break down complex questions, iteratively gather diverse evidence, and synthesize information through multiple reasoning steps. This paradigm seeks to enhance autonomy, reduce hallucinations, and improve factual accuracy in open-domain tasks.

Such deep research systems can be realized through either single-agent or multi-agent architectures. Single-agent systems rely on a single model to manage the entire process of question decomposition, retrieval, and synthesis, offering simplicity and shared context but facing limitations in handling highly specialized or multi-modal tasks. In contrast, multi-agent systems distribute these responsibilities among specialized agents, enabling modularity and potentially greater robustness. However, this collaborative design introduces additional complexity in coordination and communication, as well as higher computational costs.

Alongside these developments in agent orchestration, the nature of retrievers used in deep research has also evolved significantly. Early RAG systems relied on sparse keyword-based retrieval, later surpassed by dense retrievers employing bi-encoder architectures for semantic matching. More recent deep research systems increasingly integrate web search-based retrievers, allowing real-time access to open-domain information. Some retrievers have also been transformed into LLM-callable tools for flexible invocation. This evolution of retrievers has played a crucial role in enabling the sophisticated information-gathering processes required for deep research.

\section{Comparison of Reasoning Workflows and Agent Orchestration Strategies}

\begin{table*}[htbp]
\centering
\setlength{\tabcolsep}{5pt} 
\renewcommand{\arraystretch}{1.1} 
\scalebox{0.64}{
\begin{tabular}{
    P{2.5cm}
    P{3cm}
    P{6cm}
    P{6cm}
    P{6cm}
}
\toprule 
\rowcolor{header}
\textbf{Category} & \textbf{Sub-category} & \textbf{Strengths} & \textbf{Limitations} & \textbf{Suitable Scenarios} \\ 
\midrule

\multirow{5}{*}{\makecell[l]{\textbf{Reasoning} \\ \textbf{Workflow}}}
& Chain-based 
& One retrieval per reasoning step; low latency and token cost. Easy to cache and monitor. 
& An early wrong sub-query propagates; context grows fast on long chains. 
& Single-hop or short multi-hop QA where each intermediate fact is easy to access. \\
\cmidrule(l){2-5}

& Tree-based (ToT) 
& High recall: explores multiple branches in parallel, hedges against early errors. Transparent what-if traces. 
& Quadratic cost; tree branches require many retrieval calls. 
& Ambiguous or “multiple plausible paths” tasks (e.g., HotpotQA, legal reasoning) where missing one clue kills accuracy. \\
\cmidrule(l){2-5}

& Tree-based (MCTS) 
& Budget-aware exploration: focuses calls on promising branches; graceful anytime stopping. 
& Tuning-heavy and may converge to a suboptimal subtree. 
& Deep-search problems under tight API-call or token budgets (e.g., biomedical QA). \\
\cmidrule(l){2-5}

& Graph-based (Walk-on-Graph) 
& Efficient in explicit KG/document graphs; short reasoning paths on KGs. 
& Requires high-quality KGs; fails if graphs lack explicit edges; less flexible for open-web contexts. 
& Enterprise or domain-specific QA where a curated KG exists (e.g., product catalogs). \\
\cmidrule(l){2-5}

& Graph-based (Think-on-Graph) 
& Adaptive and verifiable; LLM updates a live evidence graph, allowing node-level citation checks and high factual accuracy. 
& Higher latency; many micro-tool calls; search space can explode without pruning. 
& Open-domain “deep research” or fact-dense synthesis tasks (e.g., BrowseComp, systematic reviews). \\
\midrule

\multirow{5}{*}{\makecell[l]{\textbf{Agent} \\ \textbf{Orchestration}}}
& Single-agent (Prompt-only) 
& Simple implementation via a ReAct loop; low resource overhead. 
& Constrained by prompt engineering and system design flexibility. 
& Prototyping demos and small-scale applications where simplicity outweighs performance. \\
\cmidrule(l){2-5}

& Single-agent (SFT) 
& Clear, well-defined RAG and reasoning patterns; higher precision than prompt-only approaches. 
& Requires large synthetic data; may overfit tool schemas, reducing out-of-domain generalization. 
& Production chatbots with stable APIs and predictable query formats (e.g., internal customer support). \\
\cmidrule(l){2-5}

& Single-agent (RL) 
& Adaptive RAG and reasoning yields high recall and accuracy; learns when to retrieve and reason. 
& Challenging to define suitable reward signals; computationally expensive to train. 
& Open-domain research or long-form QA where call costs are high and optimal stop conditions matter. \\
\cmidrule(l){2-5}

& Multi-agent (Decentralized) 
& High recall via parallel domain experts; robustness to noisy or diverse corpora. 
& High communication and consensus overhead; conflicting answers require resolution. 
& Large-scale evidence aggregation across heterogeneous sources (e.g., meta-analysis, news tracking). \\
\cmidrule(l){2-5}

& Multi-agent (Centralized/Hierarchical) 
& Budget-efficient: manager avoids duplicate searches and ensures a clear provenance chain. Scales horizontally without exponential cost growth. 
& Manager prompts or policies can become a single-point bottleneck, limiting performance. 
& Complex tasks requiring coordinated subtasks under strict API-call budgets. \\
\bottomrule
\end{tabular}}
\caption{Comparison of reasoning workflows and agent orchestration in Synergized RAG-Reasoning systems.}
\label{tab:rag_reasoning_compare}
\end{table*}

Table~\ref{tab:rag_reasoning_compare} summarizes the diverse reasoning workflows and agent orchestration strategies employed in Synergized RAG-Reasoning systems, highlighting their respective strengths, limitations, and suitable application scenarios. Reasoning workflows vary from linear chain-based approaches, which are efficient but vulnerable to error propagation, to more complex tree-based and graph-based methods that offer higher recall and transparency at the cost of increased computational overhead. Similarly, agent orchestration strategies range from single-agent setups to multi-agent systems that distribute specialized roles among agents, enhancing robustness and scalability. However, these advanced designs often introduce additional communication overhead and complexity in conflict resolution. This comparison illustrates the trade-offs inherent in choosing particular workflows or orchestration architectures and underscores the need for adaptive systems that can dynamically balance efficiency, accuracy, and resource constraints in real-world applications.

\end{document}